\documentclass{article}

\PassOptionsToPackage{numbers, compress}{natbib}

\usepackage[main, final]{neurips_2026}

\providecommand{\linenumbers}{}
\providecommand{\nolinenumbers}{}

\usepackage[utf8]{inputenc} 
\usepackage[T1]{fontenc}    
\usepackage{hyperref}       
\usepackage{url}            
\usepackage{booktabs}       
\usepackage{amsfonts}       
\usepackage{nicefrac}       
\usepackage{microtype}      
\usepackage{xcolor}         

\usepackage{tcolorbox}
\tcbuselibrary{skins, breakable} 
\usepackage{xcolor} 
\usepackage{enumitem} 
\usepackage{tcolorbox}
\usepackage{sourcecodepro}
\usepackage{float}      
\usepackage{placeins}   
\usepackage{longtable}
\usepackage{amsmath}
\usepackage{algorithm}
\usepackage{algorithmic}
\tcbuselibrary{skins}
\tcbuselibrary{skins, breakable, listings} 
\definecolor{InitialColor}{HTML}{7471C0}
\definecolor{ExplorerColor}{HTML}{EF6A6C}
\definecolor{GuardianColor}{HTML}{FDB869}
\definecolor{EvolverColor}{HTML}{A0BF43}

\title{R$^{2}$E: A Role-driven Reward Evolutionary Framework for Automated Reward Function Design}

\author{%
  \textbf{Shouhao Chang}$^{1}$\quad
  \textbf{Xuan Liu}$^{1}$\thanks{Corresponding authors.}\quad
  \textbf{Hongye Zhu}$^{1}$\quad
  \textbf{Xinning Chen}$^{1}$\quad
  \textbf{Shigeng Zhang}$^{2}$\footnotemark[1]\\[0.4em]
  \mdseries
  $^{1}$Hunan University\\
  $^{2}$Central South University
}

\begin{document}

\maketitle

\begin{abstract}
Designing effective reward functions is a critical challenge in Reinforcement Learning (RL), which traditionally requires costly manual trial-and-error. 
While Large Language Model (LLM)-based methods have shown promise in reward design, undifferentiated sampling and limited use of historical feedback often lead to insufficient exploration of the reward space and optimization instability.
To address these challenges, we propose \textbf{R$^{2}$E}, a \textbf{R}ole-driven \textbf{R}eward \textbf{E}volutionary framework for automated reward function design that structures the process through explicit role specialization.
R$^{2}$E employs multiple LLM roles with complementary objectives: an Explorer that promotes novelty to expand the search space, a Guardian that performs conservative refinements to improve training stability, and an Evolver that recombines high-performing rewards via crossover and mutation to integrate effective structures. 
A dynamic scheduling strategy coordinates these roles, progressively shifting the search from broad exploration to focused exploitation. 
To mitigate optimization instability and provide a medium for cross-role collaboration, R$^{2}$E incorporates a Global Elite Pool that retains the best-performing rewards to guide subsequent generations, thereby enabling coordinated interaction among roles and ensuring a reliable and consistent refinement process.
Extensive experiments on multiple robotic tasks demonstrate the effectiveness of our method.
Code is available at {\urlstyle{same}\url{https://github.com/ChangYuandao/R2E}}.
\end{abstract}

\section{Introduction}

Reward function design is a fundamental component in Reinforcement Learning, directly shaping agent behaviors and training stability~\cite{sutton1998reinforcement, ng1999policy}, especially in complex robotic control tasks~\cite{kober2013reinforcement,schulman2017proximal}. 
Although the field of robotic control has witnessed significant progress~\cite{hwangbo2019learning, akkaya2019solving, brohan2022rt}, achieving high-performance behaviors often necessitates intricate reward shaping.
However, manual reward engineering remains heavily dependent on domain expertise~\cite{barto2012intrinsic}, which significantly limits its scalability across diverse tasks. 
To address these challenges, Inverse Reinforcement Learning (IRL)~\cite{ng2000algorithms, ziebart2008maximum} and Preference-based Reinforcement Learning (PbRL)~\cite{wirth2017survey,christiano2017deep} have emerged as alternatives to reduce expert reliance. 
Nevertheless, they typically incur prohibitive data costs in physical environments~\cite{akrour2012april} and yield black-box representations that lack the structural transparency essential for safety and debugging~\cite{puiutta2020explainable}.

\begin{figure*}[t]
    \centering
    \includegraphics[width=\linewidth]{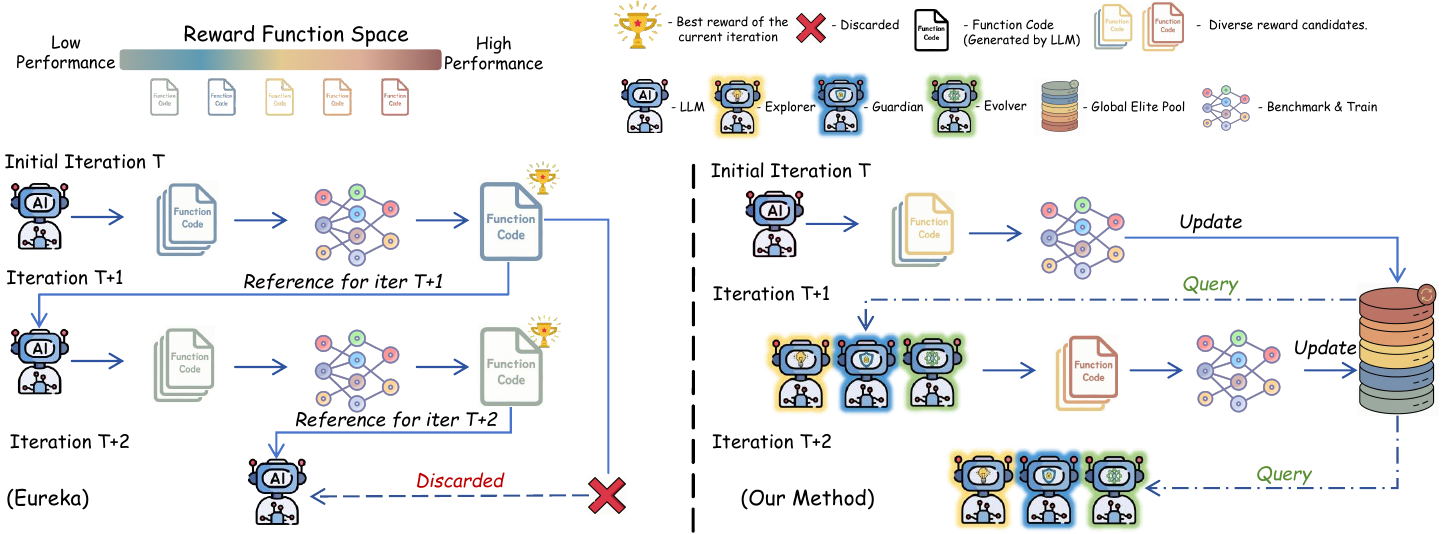}
    \caption{\textbf{Comparison of reward design paradigms.} 
    \textbf{(Eureka)} generates homogeneous rewards and lacks memory. If iteration $T+1$ regresses, the LLM references only $T+1$ for $T+2$, discarding high-performing rewards from $T$.
    \textbf{(Our Method)} populates a Global Elite Pool via initial exploration. Specialized roles query historical elites from the pool to iteratively evolve diverse reward candidates.}
    \label{fig:teaser}
\end{figure*}

Recent advances in LLM have demonstrated strong capabilities in code synthesis and logical reasoning~\cite{achiam2023gpt,chen2021evaluating}, enabling new approaches to reward function design~\cite{yulanguage,xie2024text2reward}. 
Building on these capabilities, prior work has proposed iterative frameworks that alternate between reward generation, policy training, and refinement based on feedback~\cite{ma2024eureka}.
By leveraging environmental feedback, these instruction-driven frameworks enable automated reward discovery, demonstrating substantial potential for scaling to complex, real-world RL applications~\cite{ma2024dreureka}.

Despite these advances, existing LLM-based reward generation frameworks exemplified by Eureka are still constrained by two key limitations, as illustrated in Fig.~\ref {fig:teaser}(Left).
First, reward candidates are typically produced through undifferentiated sampling under a monolithic prompting paradigm, where diversity arises mainly from stochasticity rather than a structured search process, making it difficult to explicitly balance exploration and exploitation.
Second, most iterative frameworks rely solely on the best reward and feedback from the immediate preceding iteration, lacking a persistent mechanism to preserve and reuse historically high-performing rewards.
As a result, due to the stochasticity in LLM generation, the candidates produced in the next iteration may be inferior to those from the previous iteration, leading to unstable optimization and potential performance regression.

To this end, we propose \textbf{R$^{2}$E}, a role-driven reward evolutionary framework for automated reward function design that structures the process through explicit role specialization, as illustrated in Fig.~\ref{fig:teaser}(Right).
R$^{2}$E structures the reward search process into three complementary roles: the Explorer, the Guardian, and the Evolver. 
Through a dynamic scheduling strategy, the framework adaptively adjusts the sampling allocation of each role across iterations, enabling a gradual transition from broad exploration to stable refinement. 
Furthermore, R$^{2}$E incorporates a Global Elite Pool as a persistent retention mechanism to store globally best-performing reward functions. 
This pool provides high-quality references for subsequent iterations, enabling best-so-far rewards retention and mitigating optimization instability in iterative LLM-based optimization.

The primary contributions of this work are summarized as follows:

\begin{itemize}
    \item We propose R$^{2}$E, a structured reward evolutionary framework that leverages explicit role specialization and a dynamic scheduling strategy to guide reward function design, enabling a progressive shift from broad exploration to focused exploitation.
    \item To facilitate cross-role collaboration, we introduce a Global Elite Pool. By actively retaining the highest-performing reward functions, it mitigates optimization instability and provides a stable foundation for roles to collectively refine designs.
    \item We demonstrate that R$^{2}$E consistently outperforms baselines across 14 tasks from IsaacGym and Bi-DexHands by effectively discovering high-performing reward functions for locomotion and manipulation.
\end{itemize}

\section{Related Work}

\textbf{Large Language Model for Reward Engineering.} \quad
Departing from manual engineering and data-intensive Inverse RL~\cite{abbeel2004apprenticeship}, recent research has pioneered the use of LLM for automated reward design~\cite{yulanguage, xie2024text2reward, ma2024eureka}. 
Early frameworks such as L2R~\cite{yulanguage} and Text2Reward~\cite{xie2024text2reward} established the potential of LLM but remained heavily reliant on task-specific templates, as well as few-shot prompting and human feedback.

\textbf{LLM-driven Evolutionary Reward Design.} \quad
Recent studies adopt an evolutionary paradigm to refine rewards via the iterative potential of LLM.
A landmark work in this category is Eureka~\cite{ma2024eureka}, which introduced a fully autonomous iteration cycle consisting of reward generation, training, and reflection. 
By utilizing environmental feedback to refine rewards without human intervention, Eureka demonstrated that LLM could independently discover effective reward structures. 
Subsequent research~\cite{huang2025efficient,zeng2024learning,li2025r,hazra2025revolve,gao2026rfagent} has integrated additional strategies to enhance this evolutionary process. 
For example, Revolve~\cite{hazra2025revolve} evaluates trajectories using an ELO-based scoring system, though it still incorporates costly human feedback.
RF-Agent~\cite{gao2026rfagent} regards the LLM-based reward function design as a Monte Carlo Tree Search process.
Critically, these methods (including Eureka) treat LLM as a role-agnostic stochastic sampler, which can hinder effective exploration and exploitation of high-performing outcomes across iterations.
Building upon evolutionary frameworks like Eureka, $R^{2}E$ replaces undifferentiated sampling with parallel, role-driven specialization to iteratively optimize rewards, achieving a principled balance between exploration diversity and stable performance.

\section{Problem Setting}

We study automated reward design for reinforcement learning in complex environments (e.g., physics simulations).
Similar to meta-reinforcement learning~\cite{goldie2025should}, we formulate this as a bi-level optimization problem: an outer loop searches a structured reward space generated by an LLM, while an inner-loop RL algorithm learns a policy under a fixed reward.
A task is modeled as an MDP~\cite{puterman1990markov} $\mathcal{M}=\langle \mathcal{S}, \mathcal{A}, \mathcal{P}, R, \gamma \rangle$, where $\mathcal{S}$ and $\mathcal{A}$ are the state and action spaces, $\mathcal{P}$ the transition dynamics, $R$ the reward function, and $\gamma \in [0,1]$ the discount factor.
In our framework, rewards are not predefined but generated by an LLM from a structured space $\mathcal{R}$, with $r \in \mathcal{R}$ denoting a candidate reward.

Given a reward function $r$, the corresponding policy is obtained by solving the inner-loop optimization:
\begin{equation}
\pi_r = \arg\max_\pi \mathbb{E}_{\tau \sim (\pi, \mathcal{P})}
\left[ \sum_{t=0}^{\infty} \gamma^t r(s_t, a_t) \right]
\end{equation}
where $\tau = (s_0, a_0, \ldots, s_T)$ denotes a trajectory generated by executing policy $\pi$ in the environment. 
Through this process, the agent interacts with the environment's physics $\mathcal{P}$ and learns to master complex skills, using the reward function $r$ as a guide for its behavioral improvement.
The objective of reward design is to identify an optimal reward function in the outer loop:
\begin{equation}
r^* = \arg\max_{r \in \mathcal{R}} \mathcal{J}(\pi_r)
\end{equation}
where $\mathcal{J}(\pi_r)$ denotes the expected task-level performance achieved by the policy $\pi_r$ trained under reward $r$. On IsaacGym, $\mathcal{J}$ is a task-specific performance metric; on Bi-DexHands, it is the success rate, defined on $[0,1]$. 
This task performance metric serves as a global evaluation of the reward's effectiveness in guiding the agent toward the successful completion of the target task.
This metric is used exclusively for outer-loop reward selection and is not directly observable or optimized by the inner-loop RL algorithm.

To solve this outer-loop optimization, we adopt an iterative reward evolution process. At each iteration $t$, the framework maintains a fixed-capacity Global Elite Pool $\mathcal{B}_t = \{r^{(1)}, \dots, r^{(K)}\}$ that serves as a repository for the most effective reward functions evaluated throughout the evolution. 
Conditioned on the historical rewards in $\mathcal{B}_t$, explicit role specialization is employed where multiple specialized roles including Explorer, Guardian, and Evolver collaboratively produce a set of candidate rewards:
\begin{equation}
\mathcal{R}_t = \mathcal{R}_t^{\textsc{Explorer}} \cup \mathcal{R}_t^{\textsc{Guardian}} \cup \mathcal{R}_t^{\textsc{Evolver}}
\end{equation}

characterized by complementary search objectives and dynamically adjusted sample budgets to facilitate a progressive shift from broad exploration to focused exploitation. Each candidate reward $r \in \mathcal{R}_t$ is evaluated through RL training on IsaacGym or Bi-DexHands tasks. 
If its performance $\mathcal{J}(\pi_r)$ exceeds that of the worst-performing reward in the current Global Elite Pool, it replaces that inferior candidate, ensuring the pool's quality consistently improves until the update of $\mathcal{B}_{t+1}$ is complete. After a predefined number of iterations, the final reward is selected as the best-performing element in the Global Elite Pool.

\section{Method}
\label{method}
\begin{figure*}[t]
    \centering
    \includegraphics[width=\textwidth]{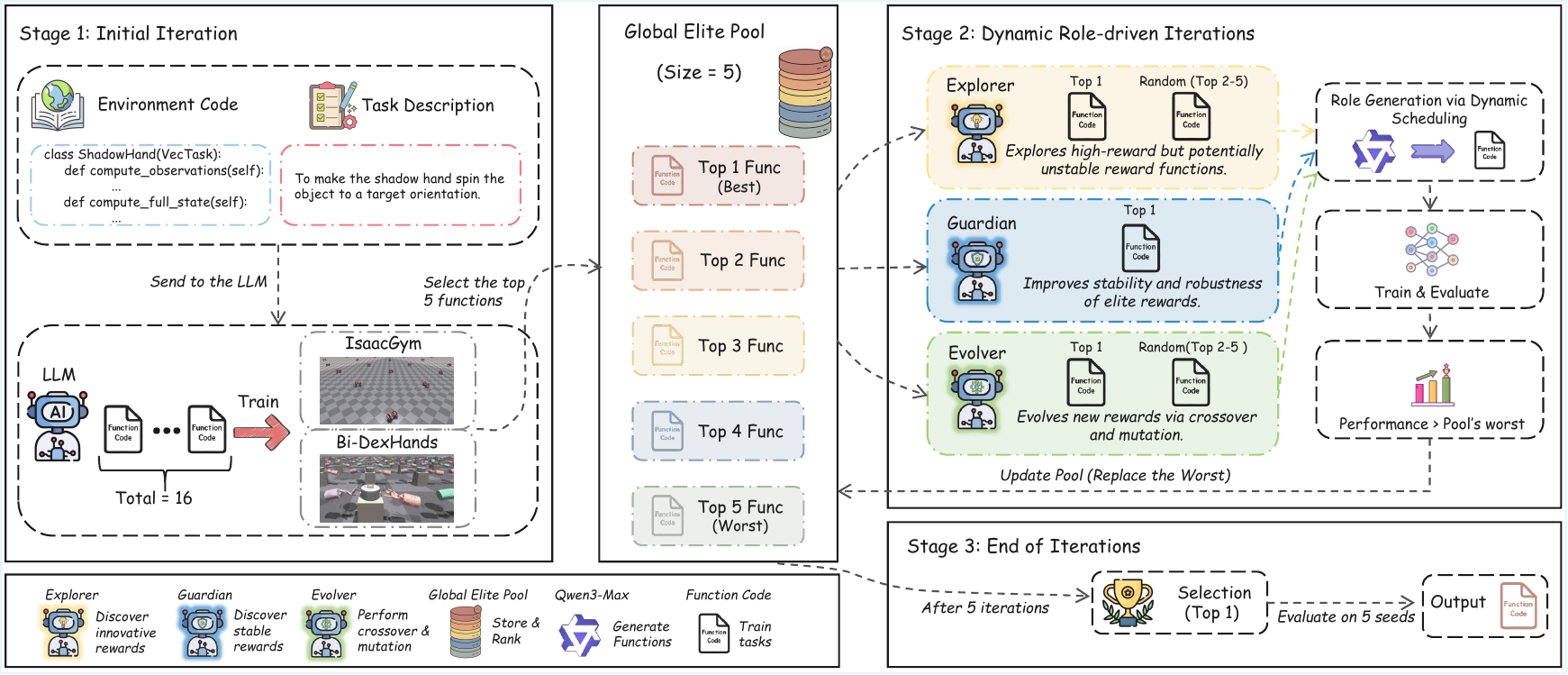}
    \caption{\textbf{Overall architecture of R$^{2}$E.} The framework operates in three main stages: 
    \textbf{(Stage 1 Initial Iteration)}: Reward candidates are generated through a broad exploration phase based on the environment code and task descriptions, and then trained on IsaacGym or Bi-DexHands tasks to initialize the Global Elite Pool with the Top 5 functions. 
    \textbf{(Stage 2 Dynamic Role-driven Iterations)}: In each iteration, a total of $N=16$ candidates are generated through dynamic scheduling by the Explorer, Guardian, and Evolver roles. After RL training and evaluation, the pool is dynamically updated by replacing the worst-performing reward.
    \textbf{(Stage 3 End of Iterations)}: After the search concludes, the Top 1 reward is selected and re-evaluated across five random seeds to produce the final output.
    \textbf{(Global Elite Pool)}: A fixed-capacity memory ($K=5$) that stores and ranks the best-performing reward functions across iterations.}
    \label{fig:overview}
\end{figure*}

In this section, we present \textbf{R$^{2}$E}, a role-driven evolutionary framework that combines multi-role collaborative reward design with a Global Elite Pool to ensure both stability and diversity in the reward search process.

\subsection{Overview}

R$^{2}$E implements a stage-wise evolutionary strategy that begins with an initial exploration phase to ensure broad coverage of the reward design space before entering role-driven evolution, as illustrated in Fig.~\ref{fig:overview}. 

In the first iteration, reward candidates are generated in an initial exploration phase to avoid prematurely restricting the search to narrow subspaces. 
This unconstrained initialization establishes sufficient structural diversity, providing a robust foundation for subsequent role-driven evolution.

Starting from the second iteration, the framework enters a role-driven phase in which the Explorer, Guardian, and Evolver collaboratively generate reward candidates conditioned on the Global Elite Pool. 
Guided by a dynamic scheduling strategy, this phase balances exploration, refinement, and the inheritance of effective reward structures.

After the evolutionary process terminates, the framework selects the top-ranked reward function from the Global Elite Pool as the final output of R$^{2}$E and re-evaluates it across multiple random seeds for robust performance estimation.

\subsection{Role-driven Reward Evolutionary Design}

Building upon the initial seed population, R$^{2}$E introduces three specialized roles to steer the evolutionary process. 
These roles operate in a coordinated manner by querying the Global Elite Pool, enabling the framework to balance exploration, conservative refinement, and the inheritance of effective reward structures.
Detailed design specifications are provided in Appendix~\ref{app:roles_engineering}.

\textbf{Explorer.} \quad
The Explorer is tasked with diversifying the reward function space by prioritizing structural creativity and unconventional reward formulations, with limited emphasis on immediate training stability.
Specifically, the Explorer queries two reward functions from the Global Elite Pool, namely the current top-performing reward and a randomly sampled remaining reward, to serve as behavioral anchors.
By analyzing the specific behaviors encouraged or overlooked by these references, the Explorer deliberately proposes new reward terms and behavioral logic that diverge from established patterns. 
This role focuses on exploring diverse reward structures and innovative shaping terms while strictly adhering to physical plausibility and numerical safety constraints.
By explicitly shifting its heuristic focus away from existing elite reward functions, the Explorer prevents the search from stagnating in local optima and uncovers novel behavioral patterns that facilitate higher performance ceilings.

\textbf{Guardian.} \quad
The Guardian focuses on enhancing the robustness and stability of reward functions through conservative refinement. 
Specifically, the Guardian queries the current best reward function from the Global Elite Pool to analyze its behavioral metrics and identify potential issues such as action oscillations or component imbalances. 
By applying stabilizing operations, including smoothing reward scales and rescaling coefficients, the Guardian generates improved variants that promote steady and predictable performance gains. 
These refinements are implemented to ensure numerical stability and magnitude balancing while preserving the core task logic of the original reward function.
Through this local optimization process, the Guardian discovers more trainable versions of established rewards and prevents performance regressions caused by noisy or unstable reward signals.

\textbf{Evolver.} \quad
The Evolver crosses over and mutates high-performing reward functions to synthesize superior offspring with enhanced behavioral characteristics. 
Specifically, the Evolver queries the current top-performing reward function and a randomly sampled candidate from the remaining members of the Global Elite Pool to serve as parent formulations. 
It identifies and merges complementary components between these diverse parents, such as combining terms that facilitate smooth motion from one with those that drive high task success from another. 
Simultaneously, the Evolver introduces controlled variations through weight recalibration, shaping adjustments, or the addition of safety constraints. 
By balancing the inheritance of successful traits with the introduction of innovative variations, the Evolver ensures that new formulations remain grounded in prior designs while promoting continual improvement.

\subsection{Global Elite Pool}

We maintain a Global Elite Pool $\mathcal{B} = \{r_1, r_2, \dots, r_K\}$ as a size-constrained repository of the best-performing reward functions discovered during the evolutionary process. 
Each $r_i \in \mathcal{B}$ represents a reward function admitted after performance evaluation. 
By setting $K=5$, we preserve diverse behavioral references while excluding lower-performing candidates that might degrade the search. 
This pool acts as a persistent memory, ensuring that successful reward structures are retained across iterations despite the inherent stochasticity of LLM generation.

\textbf{Competitive Admission Strategy.} \quad
Each candidate reward function $r$ produced by the LLM is evaluated by $\mathcal{J}(\pi_r)$ within the target environment: a task-specific performance metric on IsaacGym, and the success rate on Bi-DexHands. 
To ensure a monotonically improving boundary of reward quality, we implement a competitive admission criterion. 
A new reward function $r$ is admitted into $\mathcal{B}$ if and only if 
$\mathcal{J}(\pi_r) > \min_{r_i \in \mathcal{B}} \mathcal{J}(\pi_{r_i})$. 
Upon admission, the reward function with the lowest task performance is evicted to maintain the fixed capacity. 
This strategy transforms the pool into a dynamic filter, establishing a consistently rising performance floor that stabilizes the evolutionary trajectory.

\textbf{Inter-Role Synergy via Shared Memory.} \quad
Beyond simple storage, the Global Elite Pool functions as a central hub that facilitates synergy and interconnection among the specialized generation roles. By querying elite reward functions from the pool, the Explorer, Guardian, and Evolver transition from isolated zero-shot generation to a collaborative, guided in-context evolution.
This shared memory provides the LLM with concrete examples of best-performing reward functions and their corresponding performance metrics, serving as a structured prior that provides guidance for the search processes of different roles.
By grounding their specific heuristics in proven behavioral patterns, the roles effectively synchronize their efforts to uncover higher performance ceilings.

\subsection{Dynamic Scheduling Strategy}

We employ a dynamic scheduling strategy to manage the transition from exploration to refinement by reallocating a fixed budget of $N=16$ candidates per iteration. 
Starting from the second iteration, this strategy distributes the budget across roles to guide their search based on the Global Elite Pool.
We initially prioritize the Explorer to maximize coverage and novelty. 
As the search nears the limit $T$, the allocation shifts toward the Guardian and Evolver to transition from discovery to stability and recombination. 
This adaptive scheduling gradually shifts the focus from broad exploration to conservative refinement and inheritance, ensuring both diverse search coverage and deep optimization of reward functions.
Detailed formal definitions and the specific scheduling formulas for this reallocation are provided in Appendix~\ref{app:llm_settings}.

\begin{table*}[t]
\centering
\renewcommand{\arraystretch}{1.15}
\setlength{\tabcolsep}{6pt}
\caption{\textbf{Performance Comparison on IsaacGym Tasks.} All results report task-specific performance metrics as mean $\pm$ standard deviation across runs. \textbf{Bold} indicates the best results within each group. Avg norm score denotes the average human-normalized performance computed as $(\text{Method} - \text{Sparse})/(\text{Human} - \text{Sparse})$ for each task.}
\label{tab:Isaacgym}

\begin{small}
\resizebox{\textwidth}{!}{
\begin{tabular}{lccccccc}
\hline\hline
Method & Ant & Anymal & Quadcopter & AllegroHand & FrankaCabinet &  ShadowHand  & Avg Norm Score \\
\hline
Sparse & 1.0000{\scriptsize $\pm$0.0000} & -1.9619{\scriptsize $\pm$0.2734} & -0.3447{\scriptsize $\pm$0.2361} & 0.0619{\scriptsize $\pm$0.0415} & 0.0000{\scriptsize $\pm$0.0001} &  0.0546{\scriptsize $\pm$0.0046} & 0.0000 \\
Human  & 6.3275{\scriptsize $\pm$0.3159} & -0.0232{\scriptsize $\pm$0.0030} & -0.0906{\scriptsize $\pm$0.0095} &  11.3698{\scriptsize $\pm$0.8865}& 0.1230{\scriptsize $\pm$0.0892} & 6.9433{\scriptsize $\pm$1.3385} & 1.0000 \\
Eureka & 6.5876{\scriptsize $\pm$0.2602} & -0.0249{\scriptsize $\pm$0.0019
} & -0.0415{\scriptsize $\pm$0.0114} & 19.2377{\scriptsize $\pm$3.0125} & 0.3816{\scriptsize $\pm$0.2243} &  4.9272{\scriptsize $\pm$2.3201}& 1.4578 \\
REvolve & 4.4297{\scriptsize $\pm$1.1706} & -0.0172{\scriptsize $\pm$0.0021
} & -0.0388{\scriptsize $\pm$0.0135} & 21.9641{\scriptsize $\pm$1.3215} & 0.7224{\scriptsize $\pm$0.152} &  8.1640{\scriptsize $\pm$1.6612}& 1.8500 \\
RF-Agent & 6.7225{\scriptsize $\pm$0.2767} & -0.0154{\scriptsize $\pm$0.0053
} & -0.0274{\scriptsize $\pm$0.0028} & 20.4162{\scriptsize $\pm$2.1523} & 0.8156{\scriptsize $\pm$0.1571} &  7.9717{\scriptsize $\pm$3.5775}& 2.1512 \\
\hline
Ours & \textbf{7.2639}{\scriptsize $\pm$0.4584} & \textbf{-0.0135}{\scriptsize $\pm$0.0043} & \textbf{-0.0211}{\scriptsize $\pm$0.0076} & \textbf{23.4458}{\scriptsize $\pm$0.7224} & \textbf{1.0000}{\scriptsize $\pm$0.0000} & \textbf{8.9603}{\scriptsize $\pm$1.4171} & \textbf{2.4908} \\
\hline\hline
\end{tabular}
}
\end{small}
\end{table*}

\begin{figure*}[t]
    \centering
    \includegraphics[width=0.9\textwidth]{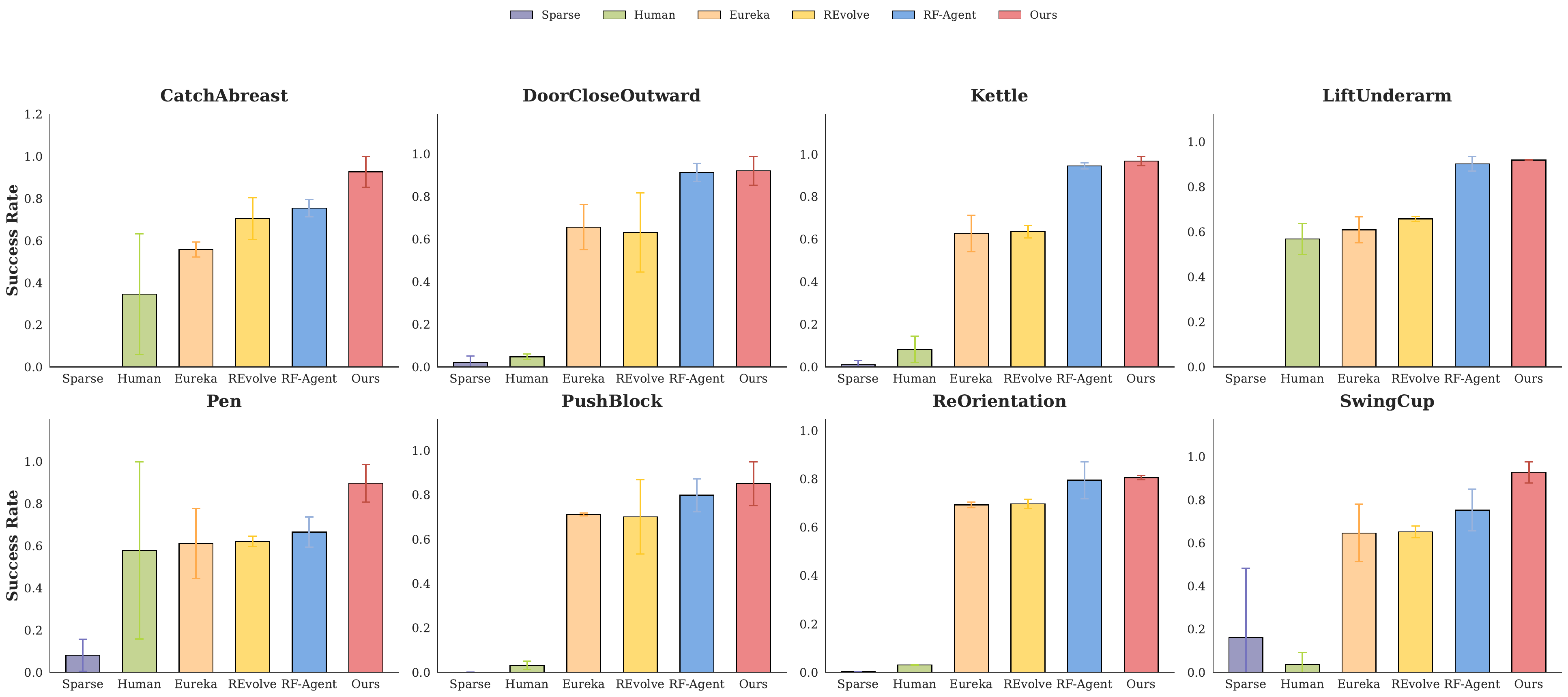}
    \caption{\textbf{Performance Comparison on Bi-DexHands Tasks.} Bars and error bars denote the mean and standard deviation of success rates over 5 random seeds, respectively. R$^2$E achieves the highest mean success rate and maintains competitive stability across all environments.}
    \label{fig:bidexhands}
\end{figure*}

\section{Experiments}

\subsection{Experimental Setup}
\label{experimental_setup}

We evaluate our framework on 14 diverse tasks from IsaacGym~\cite{makoviychuk2021isaac} and Bi-DexHands~\cite{chen2023bi}, including six representative locomotion and manipulation tasks in IsaacGym and eight complex Bi-DexHands tasks that require sophisticated dual-arm coordination,additional environment descriptions are provided in Appendix~\ref{app:environment_details}. 
Evaluation criteria are adapted to each benchmark: Bi-DexHands uses success rate because task completion is determined by a binary signal of zero or one, while IsaacGym uses task-specific performance metrics (e.g., forward speed for \emph{Ant}). 
We employ \emph{Qwen3-Max} as the LLM backbone for both our method and all baselines to ensure a fair comparison and reliable role execution during reward design.
Further implementation details are provided in Appendix~\ref{app:implementation_details}.

\subsection{Baselines}
\textbf{Sparse} replaces the dense reward with a task-level signal that is provided only when the task is successfully completed.
As a result, the reward is extremely sparse, which makes reinforcement learning significantly more challenging.

\textbf{Human} uses the dense reward functions provided by the original IsaacGym and Bi-DexHands benchmarks. These rewards were written by the researchers who designed the environments and therefore represent expert-crafted reward engineering.

\textbf{Eureka}~\cite{ma2024eureka} automatically generates dense reward functions by leveraging natural language task descriptions and environment observations. 
The method produces reward functions in batches and iteratively selects the top-performing candidate for subsequent refinement. 

\textbf{REvolve}~\cite{hazra2025revolve} utilizes a population-based approach, employing evolutionary operators to refine reward functions based on human feedback. To align with our objective of fully autonomous reward design, we utilized the auto version of REvolve, substituting human-in-the-loop input with automated environmental feedback.

\textbf{RF-Agent}~\cite{gao2026rfagent} is an automated reward function design framework that combines Large Language Models with Monte Carlo Tree Search, viewing reward design as a sequential decision-making process.

\subsection{Main Results}
\label{main_results}

We first evaluate on six representative tasks from IsaacGym, covering both locomotion tasks and manipulation tasks.
As shown in Table~\ref{tab:Isaacgym}, R$^2$E consistently achieves the best overall performance across all tasks compared with the considered baselines. 
The improvements are especially clear on manipulation tasks, demonstrating the capability of R$^2$E to handle more complex tasks characterized by high-dimensional environments and complex dynamics. 

As illustrated in Figure~\ref{fig:bidexhands}, R$^2$E achieves superior performance on the Bi-DexHands benchmark. 
We observe that our method consistently surpasses baselines, with a substantial margin in challenging tasks like \emph{CatchAbreast} and \emph{SwingCup}. 
Collectively, these results indicate that effective reward functions can be automatically designed even in more complex task environments, validating the efficacy of our method.

\begin{figure}[t]
    \centering
    \includegraphics[width=\textwidth]{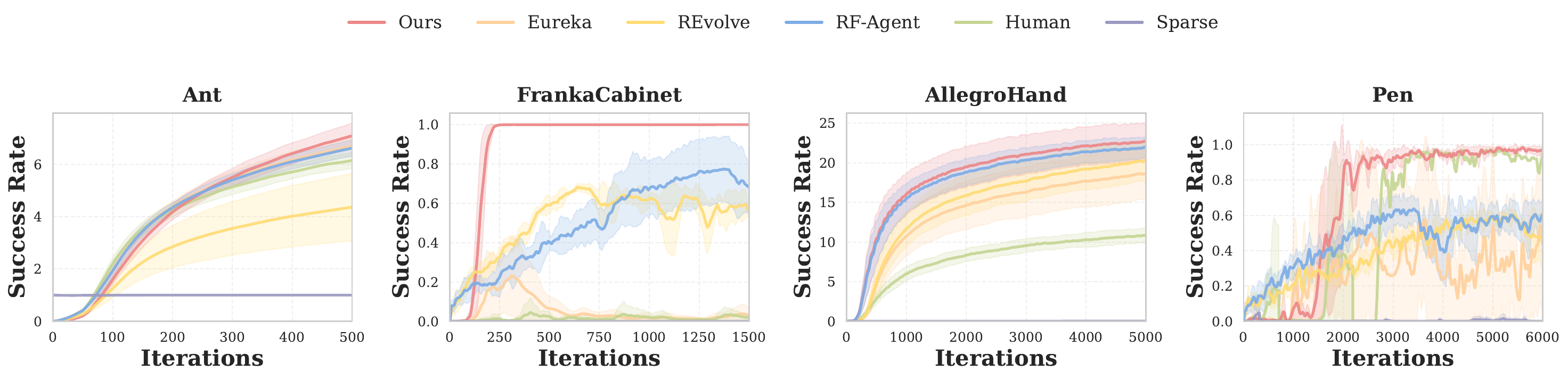}
    \caption{Training curves on representative locomotion and manipulation tasks from IsaacGym and Bi-DexHands. Solid lines and shaded regions denote the mean and standard deviation over 5 seeds. IsaacGym curves report task-specific performance metrics, and Bi-DexHands curves report success rate.} 
    \label{fig:curve}
\end{figure}

To provide a more granular analysis of the training dynamics, Figure~\ref{fig:curve} depicts the learning curves for four representative tasks covering both locomotion and manipulation. 
R$^2$E exhibits superior performance compared to baselines. 
Notably, in the \emph{FrankaCabinet} task, R$^2$E achieves rapid convergence and effectively escapes local optima to learn a viable policy.
Furthermore, in the more complex \emph{Pen} task, R$^2$E maintains high success rates with lower variance in the latter stages of training.
These experimental results demonstrate the capability of our method to design robust reward functions that facilitate stable and reliable policy learning.

\subsection{Ablation Studies}

\begin{figure}[t]
    \centering
    \begin{minipage}{0.48\textwidth}
        \centering
        \includegraphics[width=\linewidth]{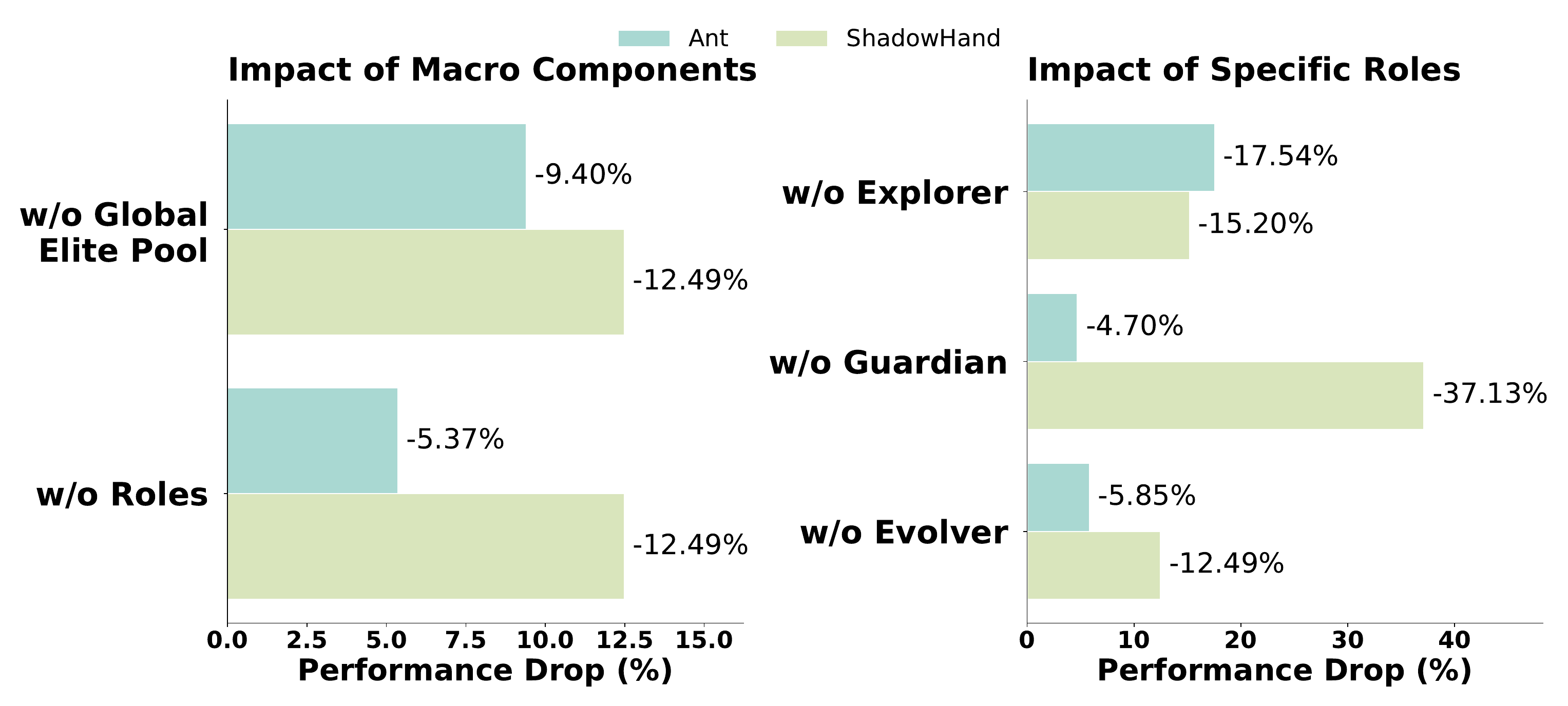}
        \caption{Ablation study on \emph{Ant} and 
        \emph{ShadowHand} tasks. The charts display the performance drop (\%) relative to the full method. \textbf{Left:} Ablation of macro components. \textbf{Right:} Ablation of specific roles. The length of the bars indicates the importance of each component.}
        \label{fig:ablation}
    \end{minipage}\hfill
    \begin{minipage}{0.48\textwidth}
        \centering
        \includegraphics[width=\linewidth]{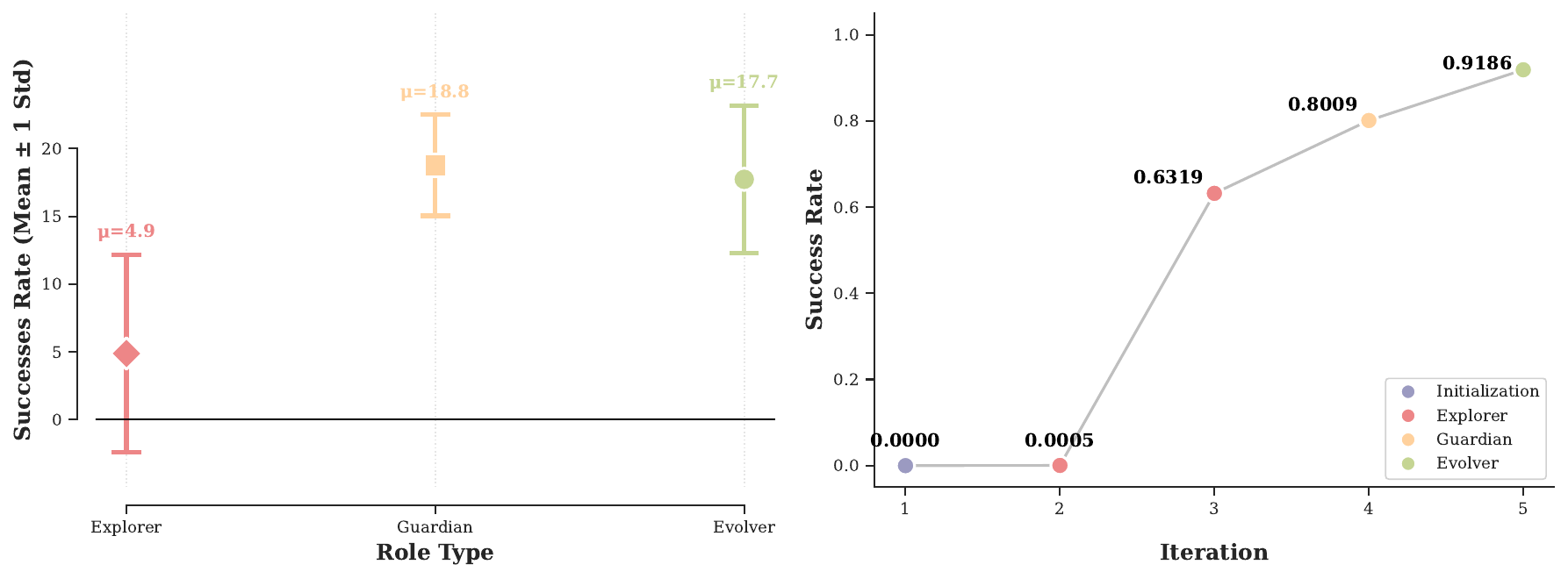}
        \caption{Analysis of Role Distinctiveness and Synergistic Evolution. 
        \textbf{Left:} Validates the statistical characteristics of the roles, where $\mu$ denotes the mean reward.
        \textbf{Right:} Illustrates the synergistic mechanism in a representative evolutionary trajectory.}
        \label{fig:analyse}
    \end{minipage}
\end{figure}

To understand the contribution of each component in our framework, we conduct ablation experiments on \emph{Ant} (locomotion) and \emph{ShadowHand} (manipulation).
The results are illustrated in Figure~\ref{fig:ablation}, with a detailed analysis provided below.

\textbf{w/o Roles \& Global Elite Pool.} \quad 
Removing role specialization causes clear performance degradation, indicating that role-driven diversity effectively expands the search space.
Similarly, ablating the Global Elite Pool by restricting reference functions to only the best result from the previous iteration severely destabilizes optimization. 
Together, these results show that both diversified generation patterns and the preservation of globally superior references are essential for discovering high-performing reward functions.

\textbf{w/o Individual Roles.} \quad
To evaluate the contribution of each role, we conducted ablation experiments by removing the Explorer, Guardian, or Evolver and redistributing its responsibilities among the remaining modules.
The results show that the removal of any single role leads to significant performance degradation. Notably, the removal of the Guardian causes a substantial drop in performance on the \emph{ShadowHand} task, reflecting the difficulty of reward design in complex tasks and highlighting the Guardian’s critical role in stabilizing the learning process.
These findings demonstrate that each module is indispensable in the iterative optimization process, as they collectively provide complementary mechanisms of exploration, refinement, and evolution to sustain continuous improvement.

\textbf{Impact of Dynamic Scheduling Strategy.} \quad
We evaluate scheduling effectiveness by comparing our Dynamic Scheduling Strategy with a fixed baseline that uses 5 Explorers, 5 Guardians, and 6 Evolvers per iteration.
Conversely, our dynamic approach allocates more Explorers early to discover diverse reward structures and shifts toward Guardians and Evolvers later for precise optimization. 
Table~\ref{tab:scheduling} demonstrates that dynamic scheduling consistently improves the task-specific performance metrics on the IsaacGym tasks and the success rate on \emph{Pen}. 
This confirms that transitioning from broad exploration to localized refinement enables fine-grained optimization of promising candidates and ensures robust final performance.

\begin{table}[h]
    \centering
    \caption{Impact of Scheduling Strategy on task performance.}
    \resizebox{\linewidth}{!}{
    \begin{tabular}{lcccc} 
        \toprule
        \textbf{Scheduling Strategy} & \textbf{Ant} & \textbf{FrankaCabinet} & \textbf{AllegroHand} & \textbf{Pen} \\
        \midrule
        Fixed (5:5:6 Allocation) & $6.5833 \pm 0.3008$ & $0.8157 \pm 0.0206$ & $20.7247 \pm 1.9726$ & $0.7277 \pm 0.1685$ \\
        Dynamic (Ours) & $\mathbf{7.2639 \pm 0.4584}$ & $\mathbf{1.0000 \pm 0.0000}$ & $\mathbf{23.4458 \pm 0.7224}$ & $\mathbf{0.8967 \pm 0.0891}$ \\
        \bottomrule
    \end{tabular}
    }
    \label{tab:scheduling}
\end{table}

\textbf{Impact of Global Elite Pool Size ($K$).} \quad
We further investigate the sensitivity of $R^2E$ to the capacity of the Global Elite Pool by varying the size $K \in \{3, 5, 8, 12, 16\}$ across four representative tasks. 
As shown in Table~\ref{tab:pool_size}, setting $K=5$ consistently yields the optimal performance. 
A smaller pool size restricts the structural diversity of the retained reward functions, often causing the search process to prematurely converge to local optima. 
Conversely, employing a larger pool under a fixed generation budget lowers the admission threshold of the elite pool; it forces the framework to more frequently sample sub-optimal parents for crossover and mutation, which ultimately degrades both search efficiency and final performance.

\begin{table}[h]
    \centering
    \caption{Ablation study on the size of the Global Elite Pool ($K$).}
    \resizebox{\textwidth}{!}{
    \begin{tabular}{lccccc}
        \toprule
        \textbf{Task} & $K=3$ & $K=5$ \textbf{(Ours)} & $K=8$ & $K=12$ & $K=16$ \\
        \midrule
        Ant & $6.8632 \pm 0.2850$ & $\mathbf{7.2639 \pm 0.4584}$ & $6.7844 \pm 0.8383$ & $5.5527 \pm 0.4833$ & $3.8366 \pm 3.0366$ \\
        FrankaCabinet & $0.8621 \pm 0.1026$ & $\mathbf{1.0000 \pm 0.0000}$ & $0.7425 \pm 0.0156$ & $0.7627 \pm 0.1036$ & $0.5163 \pm 0.1673$ \\
        AllegroHand & $23.2724 \pm 1.6273$ & $\mathbf{23.4458 \pm 0.7224}$ & $22.7741 \pm 1.7614$ & $21.6571 \pm 1.6136$ & $18.1642 \pm 2.6115$ \\
        Pen & $0.8201 \pm 0.0617$ & $\mathbf{0.8967 \pm 0.0891}$ & $0.8013 \pm 0.0968$ & $0.7314 \pm 0.0916$ & $0.5614 \pm 0.2901$ \\
        \bottomrule
    \end{tabular}
    }
    \label{tab:pool_size}
\end{table}
\subsection{Qualitative Analysis \& Case Studies}

To provide a deeper understanding of the internal mechanisms of R$^{2}$E, we conduct a comprehensive qualitative analysis to answer three key research questions: 
\textbf{Q1.} Do the Explorer, Guardian, and Evolver roles exhibit distinct statistical characteristics consistent with their semantic definitions? 
\textbf{Q2.} How do these diverse roles collaborate to drive the evolutionary trajectory of the maximum reward? 
\textbf{Q3.} How does performance scale with the search horizon and the number of candidates in each iteration?

\textbf{Analysis of Role Distinctiveness (Q1).} \quad
Taking the \emph{AllegroHand} task as a representative example, we analyze the statistical behaviors of the three roles to verify if they align with their semantic designs. 
As illustrated in Figure~\ref{fig:analyse} (Left), the Explorer exhibits a lower mean reward ($\mu=4.9$) but the highest variance, reflecting its aggressive search for novel reward structures that prioritizes diversity over immediate stability. 
In sharp contrast, the Guardian achieves the highest mean reward ($\mu=18.8$) with minimal variance, confirming its role in performing conservative refinements to ensure reliable and consistent policy training. 
The Evolver demonstrates an intermediate statistical profile because it inherits high-performing structures from elite parents while introducing controlled variations via crossover and mutation, resulting in stability and average performance ($\mu=17.7$) that fall between those of the Explorer and the Guardian.
These distinct statistical signatures validate that the LLM effectively understands its assigned roles, demonstrating that our framework successfully transforms undifferentiated sampling into a structured search process. 
By ensuring that each specialized role performs its unique and complementary function, R$^{2}$E establishes a more robust paradigm that effectively expands the reward function search space.

\textbf{Synergy and Evolutionary Trajectory (Q2).} \quad
To better illustrate the coordination mechanism among roles during the evolution, we select a highly representative case study—the \emph{LiftUnderarm} task, as visualized in Figure~\ref{fig:analyse} (Right). 
In this task, the initial exploration phase without role assignment yields zero performance, highlighting the formidable challenges of reward design in the early stages. 
Nevertheless, these initial attempts are preserved in the Global Elite Pool. 
By analyzing these failed reward structures, the Explorer is able to identify ineffective patterns and innovatively design functional reward structures, thereby effectively expanding the search space. Subsequently, the Guardian refines and stabilizes the reward logic, leading to a significant performance leap. 
Finally, the Evolver leverages the strengths of elite candidates from the global history pool, combining the Explorer’s exploration capability with the Guardian’s stability to derive the globally optimal reward function. 
This evolutionary trajectory clearly demonstrates how the distinct roles collaborate to guide the search process from broad exploration to precise exploitation. 
Detailed results in Appendix~\ref{case_example}.

\textbf{Search Horizon and Sampling Budget (Q3).} \quad
We next examine whether the reported gains depend on a particular search budget. 
All other settings follow the main experiments. 
On \emph{Ant}, \emph{AllegroHand}, and \emph{Pen}, we first fix the number of candidates in each iteration at $N=16$ and vary only the number of iterations $T$. 
\emph{Ant} and \emph{AllegroHand} are measured by task-specific performance metrics, and \emph{Pen} by success rate. 
As shown in Table~\ref{tab:horizon}, performance improves on every task as $T$ increases from 1 to 7. 
Most of the gain appears by $T=5$; extending the search to $T=7$ brings a smaller additional improvement. 
We then fix $T=5$ and vary only $N$. 
From the second iteration, the same dynamic schedule reallocates this budget in the order (Explorer, Guardian, Evolver). 
The four subsequent iterations use $(3,0,1)\rightarrow(2,1,1)\rightarrow(1,1,2)\rightarrow(0,2,2)$ when $N=4$, $(6,1,1)\rightarrow(4,2,2)\rightarrow(2,3,3)\rightarrow(0,4,4)$ when $N=8$, $(12,2,2)\rightarrow(8,4,4)\rightarrow(4,6,6)\rightarrow(0,8,8)$ when $N=16$, and $(18,3,3)\rightarrow(12,6,6)\rightarrow(6,9,9)\rightarrow(0,12,12)$ when $N=24$. 
Table~\ref{tab:budget} shows that sampling more candidates in each iteration also improves all three tasks. 
Increasing either the horizon or the number of candidates therefore strengthens the search of R$^{2}$E. 
We set $N=16$ and $T=5$ in the main comparison so that the evaluation budget matches that of the baselines. 
Adapting $T$ and $N$ during search, rather than specifying them in advance, remains future work.

\begin{table}[t]
    \centering
    \caption{Performance under different search horizons with $N=16$. \textbf{Bold} indicates the highest value in each task. $T=5$ is the setting used in the main experiments.}
    \label{tab:horizon}
    \resizebox{\linewidth}{!}{
    \begin{tabular}{lcccc}
        \toprule
        \textbf{Iterations} ($T$) & \textbf{Total candidates} ($T \times N$) & \textbf{Ant} & \textbf{AllegroHand} & \textbf{Pen} \\
        \midrule
        1 & 16 & $4.8317 \pm 0.3143$ & $16.5829 \pm 1.6192$ & $0.3873 \pm 0.1354$ \\
        3 & 48 & $5.7638 \pm 0.1865$ & $20.6347 \pm 1.0519$ & $0.6516 \pm 0.1017$ \\
        5 & 80 & $7.2639 \pm 0.4584$ & $23.4458 \pm 0.7224$ & $0.8967 \pm 0.0891$ \\
        7 & 112 & $\mathbf{7.6821 \pm 0.6817}$ & $\mathbf{24.0685 \pm 0.4739}$ & $\mathbf{0.9166 \pm 0.0539}$ \\
        \bottomrule
    \end{tabular}
    }
\end{table}

\begin{table}[t]
    \centering
    \caption{Performance under different numbers of candidates in each iteration with $T=5$. \textbf{Bold} indicates the highest value in each task. $N=16$ is the setting used in the main experiments.}
    \label{tab:budget}
    \resizebox{\linewidth}{!}{
    \begin{tabular}{lcccc}
        \toprule
        \textbf{Budget} ($N$) & \textbf{Total candidates} ($5 \times N$) & \textbf{Ant} & \textbf{AllegroHand} & \textbf{Pen} \\
        \midrule
        4 & 20 & $6.0789 \pm 0.6078$ & $19.0723 \pm 0.4367$ & $0.6049 \pm 0.1033$ \\
        8 & 40 & $6.5672 \pm 1.1067$ & $20.9287 \pm 0.8713$ & $0.7549 \pm 0.0376$ \\
        16 & 80 & $7.2639 \pm 0.4584$ & $23.4458 \pm 0.7224$ & $0.8967 \pm 0.0891$ \\
        24 & 120 & $\mathbf{7.9834 \pm 0.2569}$ & $\mathbf{24.4028 \pm 0.3644}$ & $\mathbf{0.9534 \pm 0.0135}$ \\
        \bottomrule
    \end{tabular}
    }
\end{table}

\section{Conclusion}

We propose R$^{2}$E, a role-driven reward evolutionary framework that transforms LLM-based reward design from undifferentiated sampling into a structured, collaborative search process. 
By orchestrating the Explorer, Guardian, and Evolver roles alongside a Global Elite Pool, R$^{2}$E explicitly balances broad exploration of the reward function space with the conservative refinement required for training stability.
This architecture effectively mitigates optimization instability by preserving globally best-performing candidates as persistent references, maintaining a stable evolutionary trajectory across iterations. 
Experiments on IsaacGym and Bi-DexHands demonstrate that R$^{2}$E consistently outperforms baselines, especially in complex manipulation tasks.
More importantly, R$^{2}$E pioneers a more effective paradigm for autonomous reward design.
By assigning distinct role characteristics to the LLM, our framework structurally improves reward functions during the evolutionary process, enabling the search to effectively escape local optima and discover reward functions of significantly higher quality.

Beyond the present framework, an important question is how to retain the experience contained in failed reward functions and use it together with the experience of successful ones. 
This memory should persist across iterations, so that later rounds can avoid previously ineffective structures while continuing to build on rewards that have already proved useful. 
A related question is how to decide when evolution should stop. 
Both the joint use of failure and success across iterations and a principled stopping decision merit further attention in automated reward design.

\begin{ack}
This work was supported by the National Key Research and Development Program of China (No. 2025YB3003500) and the National Natural Science Foundation of China (No. U25A20437, 62672521, 62372473).
\end{ack}

\clearpage
\bibliographystyle{plainnat}
\bibliography{references}

@book{sutton1998reinforcement,
  title={Reinforcement learning: An introduction},
  author={Sutton, Richard S and Barto, Andrew G and others},
  volume={1},
  number={1},
  year={1998},
  publisher={MIT press Cambridge}
}

@inproceedings{ng1999policy,
  title={Policy invariance under reward transformations: Theory and application to reward shaping},
  author={Ng, Andrew Y and Harada, Daishi and Russell, Stuart},
  booktitle={Icml},
  volume={99},
  pages={278--287},
  year={1999},
  organization={Citeseer}
}

@article{kober2013reinforcement,
  title={Reinforcement learning in robotics: A survey},
  author={Kober, Jens and Bagnell, J Andrew and Peters, Jan},
  journal={The International Journal of Robotics Research},
  volume={32},
  number={11},
  pages={1238--1274},
  year={2013},
  publisher={SAGE Publications Sage UK: London, England}
}

@article{schulman2017proximal,
  title={Proximal policy optimization algorithms},
  author={Schulman, John and Wolski, Filip and Dhariwal, Prafulla and Radford, Alec and Klimov, Oleg},
  journal={arXiv preprint arXiv:1707.06347},
  year={2017}
}

@article{hwangbo2019learning,
  title={Learning agile and dynamic motor skills for legged robots},
  author={Hwangbo, Jemin and Lee, Joonho and Dosovitskiy, Alexey and Bellicoso, Dario and Tsounis, Vassilios and Koltun, Vladlen and Hutter, Marco},
  journal={Science Robotics},
  volume={4},
  number={26},
  pages={eaau5872},
  year={2019},
  publisher={American Association for the Advancement of Science}
}

@article{akkaya2019solving,
  title={Solving rubik's cube with a robot hand},
  author={Akkaya, Ilge and Andrychowicz, Marcin and Chociej, Maciek and Litwin, Mateusz and McGrew, Bob and Petron, Arthur and Paino, Alex and Plappert, Matthias and Powell, Glenn and Ribas, Raphael and others},
  journal={arXiv preprint arXiv:1910.07113},
  year={2019}
}

@article{goldie2025should,
  title={How Should We Meta-Learn Reinforcement Learning Algorithms?},
  author={Goldie, Alexander David and Wang, Zilin and Cohen, Jaron and Foerster, Jakob Nicolaus and Whiteson, Shimon},
  journal={arXiv preprint arXiv:2507.17668},
  year={2025}
}

@article{brohan2022rt,
  title={Rt-1: Robotics transformer for real-world control at scale},
  author={Brohan, Anthony and Brown, Noah and Carbajal, Justice and Chebotar, Yevgen and Dabis, Joseph and Finn, Chelsea and Gopalakrishnan, Keerthana and Hausman, Karol and Herzog, Alex and Hsu, Jasmine and others},
  journal={arXiv preprint arXiv:2212.06817},
  year={2022}
}

@incollection{barto2012intrinsic,
  title={Intrinsic motivation and reinforcement learning},
  author={Barto, Andrew G},
  booktitle={Intrinsically motivated learning in natural and artificial systems},
  pages={17--47},
  year={2012},
  publisher={Springer}
}

@inproceedings{ma2024eureka,
title={Eureka: Human-Level Reward Design via Coding Large Language Models},
author={Yecheng Jason Ma and William Liang and Guanzhi Wang and De-An Huang and Osbert Bastani and Dinesh Jayaraman and Yuke Zhu and Linxi Fan and Anima Anandkumar},
booktitle={The Twelfth International Conference on Learning Representations},
year={2024}
}

@inproceedings{yulanguage,
  author       = {Wenhao Yu and
                  Nimrod Gileadi and
                  Chuyuan Fu and
                  Sean Kirmani and
                  Kuang{-}Huei Lee and
                  Montserrat Gonzalez Arenas and
                  Hao{-}Tien Lewis Chiang and
                  Tom Erez and
                  Leonard Hasenclever and
                  Jan Humplik and
                  Brian Ichter and
                  Ted Xiao and
                  Peng Xu and
                  Andy Zeng and
                  Tingnan Zhang and
                  Nicolas Heess and
                  Dorsa Sadigh and
                  Jie Tan and
                  Yuval Tassa and
                  Fei Xia},
  title        = {Language to Rewards for Robotic Skill Synthesis},
  booktitle    = {Conference on Robot Learning, CoRL 2023, 6-9 November 2023, Atlanta,
                  GA, {USA}},
  volume       = {229},
  pages        = {374--404},
  year         = {2023},
}

@inproceedings{ng2000algorithms,
  author       = {Andrew Y. Ng and
                  Stuart Russell},
  title        = {Algorithms for Inverse Reinforcement Learning},
  booktitle    = {Proceedings of the Seventeenth International Conference on Machine
                  Learning {(ICML} 2000), Stanford University, Stanford, CA, USA, June
                  29 - July 2, 2000},
  pages        = {663--670},
  year         = {2000},
}

@inproceedings{ziebart2008maximum,
  title={Maximum entropy inverse reinforcement learning.},
  author={Ziebart, Brian D and Maas, Andrew L and Bagnell, J Andrew and Dey, Anind K and others},
  booktitle={Aaai},
  volume={8},
  pages={1433--1438},
  year={2008},
  organization={Chicago, IL, USA}
}

@article{christiano2017deep,
  title={Deep reinforcement learning from human preferences},
  author={Christiano, Paul F and Leike, Jan and Brown, Tom and Martic, Miljan and Legg, Shane and Amodei, Dario},
  journal={Advances in neural information processing systems},
  volume={30},
  year={2017}
}

@article{wirth2017survey,
  title={A survey of preference-based reinforcement learning methods},
  author={Wirth, Christian and Akrour, Riad and Neumann, Gerhard and F{\"u}rnkranz, Johannes},
  journal={Journal of Machine Learning Research},
  volume={18},
  number={136},
  pages={1--46},
  year={2017}
}

@inproceedings{akrour2012april,
  title={April: Active preference learning-based reinforcement learning},
  author={Akrour, Riad and Schoenauer, Marc and Sebag, Mich{\`e}le},
  booktitle={Joint European conference on machine learning and knowledge discovery in databases},
  pages={116--131},
  year={2012},
  organization={Springer}
}

@inproceedings{puiutta2020explainable,
  title={Explainable reinforcement learning: A survey},
  author={Puiutta, Erika and Veith, Eric MSP},
  booktitle={International cross-domain conference for machine learning and knowledge extraction},
  pages={77--95},
  year={2020},
  organization={Springer}
}

@article{achiam2023gpt,
  title={Gpt-4 technical report},
  author={Achiam, Josh and Adler, Steven and Agarwal, Sandhini and Ahmad, Lama and Akkaya, Ilge and Aleman, Florencia Leoni and Almeida, Diogo and Altenschmidt, Janko and Altman, Sam and Anadkat, Shyamal and others},
  journal={arXiv preprint arXiv:2303.08774},
  year={2023}
}

@article{chen2021evaluating,
  title={Evaluating large language models trained on code},
  author={Chen, Mark},
  journal={arXiv preprint arXiv:2107.03374},
  year={2021}
}

@inproceedings{xie2024text2reward,
  title={Text2Reward: Reward Shaping with Language Models for Reinforcement Learning},
  author={Xie, Tianbao and Zhao, Siheng and Wu, Chen Henry and Liu, Yitao and Luo, Qian and Zhong, Victor and Yang, Yanchao and Yu, Tao},
  booktitle={ICLR},
  year={2024}
}

@inproceedings{ma2024dreureka,
  author       = {Yecheng Jason Ma and
                  William Liang and
                  Hung{-}Ju Wang and
                  Yuke Zhu and
                  Linxi Fan and
                  Osbert Bastani and
                  Dinesh Jayaraman},
  title        = {DrEureka: Language Model Guided Sim-To-Real Transfer},
  booktitle    = {Robotics: Science and Systems XX, Delft, The Netherlands, July 15-19,
                  2024},
  year         = {2024},

}

@inproceedings{abbeel2004apprenticeship,
  title={Apprenticeship learning via inverse reinforcement learning},
  author={Abbeel, Pieter and Ng, Andrew Y},
  booktitle={Proceedings of the twenty-first international conference on Machine learning},
  pages={1},
  year={2004}
}

@inproceedings{zeng2024learning,
  title={Learning Reward for Robot Skills Using Large Language Models via Self-Alignment},
  author={Zeng, Yuwei and Mu, Yao and Shao, Lin},
  booktitle={International Conference on Machine Learning},
  pages={58366--58386},
  year={2024},
  organization={PMLR}
}

@inproceedings{huang2025efficient,
  title={Efficient language-instructed skill acquisition via reward-policy co-evolution},
  author={Huang, Changxin and Chang, Yanbin and Lin, Junfan and Liang, Junyang and Zeng, Runhao and Li, Jianqiang},
  booktitle={Proceedings of the AAAI Conference on Artificial Intelligence},
  volume={39},
  number={14},
  pages={14576--14584},
  year={2025}
}

@article{gao2026rfagent,
  title={RF-Agent: Automated Reward Function Design via Language Agent Tree Search},
  author={Gao, N and Zhang, X and Jiang, X and others},
  journal={arXiv preprint arXiv:2602.23876},
  year={2026}
}

@inproceedings{hazra2025revolve,
title={{RE}volve: Reward Evolution with Large Language Models using Human Feedback},
author={RISHI HAZRA and Alkis Sygkounas and Andreas Persson and Amy Loutfi and Pedro Zuidberg Dos Martires},
booktitle={The Thirteenth International Conference on Learning Representations},
year={2025},
}

@inproceedings{li2025r,
title={R*: Efficient Reward Design via Reward Structure Evolution and Parameter Alignment Optimization with Large Language Models},
author={Pengyi Li and Jianye HAO and Hongyao Tang and Yifu Yuan and Jinbin Qiao and Zibin Dong and YAN ZHENG},
booktitle={Forty-second International Conference on Machine Learning},
year={2025}
}

@article{puterman1990markov,
  title={Markov decision processes},
  author={Puterman, Martin L},
  journal={Handbooks in operations research and management science},
  volume={2},
  pages={331--434},
  year={1990},
  publisher={Elsevier}
}

@article{makoviychuk2021isaac,
  title={Isaac gym: High performance gpu-based physics simulation for robot learning},
  author={Makoviychuk, Viktor and Wawrzyniak, Lukasz and Guo, Yunrong and Lu, Michelle and Storey, Kier and Macklin, Miles and Hoeller, David and Rudin, Nikita and Allshire, Arthur and Handa, Ankur and others},
  journal={arXiv preprint arXiv:2108.10470},
  year={2021}
}

@article{chen2023bi,
  title={Bi-dexhands: Towards human-level bimanual dexterous manipulation},
  author={Chen, Yuanpei and Geng, Yiran and Zhong, Fangwei and Ji, Jiaming and Jiang, Jiechuang and Lu, Zongqing and Dong, Hao and Yang, Yaodong},
  journal={IEEE Transactions on Pattern Analysis and Machine Intelligence},
  volume={46},
  number={5},
  pages={2804--2818},
  year={2023},
  publisher={IEEE}
}

@article{makoviichuk2021rl,
  title={rl-games: A high-performance framework for reinforcement learning},
  author={Makoviichuk, Denys and Makoviychuk, Viktor},
  journal={GitHub repository, May},
  year={2021}
}

\clearpage
\appendix
\section{Roles Engineering \& Templates}
\label{app:roles_engineering}

In this section, we provide the detailed prompt templates used for R$^{2}$E. We leverage a chat-based interaction format consisting of \textbf{System}, \textbf{User}, and \textbf{Assistant} messages to guide the LLM.

\subsection{Common Interface}
\label{app:common_interface}

To maintain a controlled experimental environment and ensure an identical informational baseline, the \textbf{User} message is kept strictly invariant across all assigned roles and throughout every iteration of the framework. This design choice ensures that the core task definition and environmental context delivered to the LLM remain homogeneous, regardless of the specific role-playing constraints or the current stage of the process. By employing a standardized template, we guarantee that the LLM parses the observation space and task goals from a constant source of truth, thereby isolating role-specific behavioral variances from the fundamental understanding of the task objectives.

\nolinenumbers  
\begin{tcolorbox}[
    breakable,          
    enhanced,            
    use color stack,     
    title=Common User Template,
    attach boxed title to top left={xshift=3mm, yshift=-3mm, yshifttext=-1mm},
    colback=gray!2,
    colframe=darkgray,
    colbacktitle=white,
    coltitle=darkgray,
    fonttitle=\bfseries\small,
    boxed title style={size=small, sharp corners, colframe=darkgray},
    fontupper=\fontsize{7pt}{10pt}\selectfont\ttfamily,
    arc=1mm,
    boxrule=0.5pt,
    left=4mm, right=4mm, top=5mm, bottom=2mm
]
The Python environment is \{task\_obs\_code\_string\}. Write a reward function for the following task: \{task\_description\}.
\end{tcolorbox}
\linenumbers  

\subsection{Initialization }
\label{app:initialization}

During the initial iteration, R$^{2}$E operates in an unconstrained manner without role-specific priors to facilitate a comprehensive exploration of the reward space. By prompting the LLM as a generalist reward engineer, we prioritize broad coverage of the potential reward space over specific behavioral alignment. This unconstrained phase is critical for generating a diverse set of reward candidates, which serves to seed the Global Elite Pool with a sufficiently varied population for subsequent evolutionary refinement.

\nolinenumbers  
\begin{tcolorbox}[
    breakable,          
    enhanced,            
    use color stack,     
    title=Initial System Template,
    attach boxed title to top left={xshift=3mm, yshift=-3mm, yshifttext=-1mm},
    colback=gray!2,
    colframe=InitialColor,
    colbacktitle=white,
    coltitle=InitialColor,
    fonttitle=\bfseries\small,
    boxed title style={size=small, sharp corners, colframe=InitialColor},
    fontupper=\fontsize{7pt}{10pt}\selectfont\ttfamily,
    arc=1mm,
    boxrule=0.5pt,
    left=4mm, right=4mm, top=5mm, bottom=2mm
]
You are a reward engineer trying to write reward functions to solve reinforcement learning tasks as effectively as possible. \\
Your goal is to write a free exploration reward function for the environment that will help the agent learn the task described in text. \\
Your reward function should use useful variables from the environment as inputs. As an example, the reward function signature can be: \{task\_reward\_signature\_string\} \\
Since the reward function will be decorated with @torch.jit.script, please make sure that the code is compatible with TorchScript (e.g., use torch tensor instead of numpy array). \\
Make sure any new tensor or variable you introduce is on the same device as the input tensors. 

\vspace{2mm}
The output of the reward function should consist of two items: \\
\hspace*{4mm}(1) the total reward, \\
\hspace*{4mm}(2) a dictionary of each individual reward component. \\
The code output should be formatted as a Python code string: "\texttt{\`}\texttt{\`}\texttt{\`}python ...\texttt{\`}\texttt{\`}\texttt{\`}".

\vspace{2mm}
Some helpful tips for writing the reward function code: \\
\hspace*{4mm}(1) You may find it helpful to normalize the reward to a fixed range by applying transformations like torch.exp to the overall reward or its components \\
\hspace*{4mm}(2) If you choose to transform a reward component, then you must also introduce a temperature parameter inside the transformation function; this parameter must be a named variable in the reward function and it must not be an input variable. Each transformed reward component should have its own temperature variable \\
\hspace*{4mm}(3) Make sure the type of each input variable is correctly specified; a float input variable should not be specified as torch.Tensor \\
\hspace*{4mm}(4) Most importantly, the reward code\texttt{\'}s input variables must contain only attributes of the provided environment class definition (namely, variables that have prefix self). Under no circumstance can you introduce new input variables.
\end{tcolorbox}
\linenumbers  

\subsection{Role-Driven Evolution}

From the second iteration, R$^{2}$E transitions to a structured evolutionary phase featuring three roles: Explorer, Guardian, and Evolver. All roles share an identical User Message to ensure task consistency. To optimize efficiency, their prompting structures are differentiated: the Explorer and Guardian utilize both System and Assistant messages (with elite rewards as exemplars), while the Evolver is streamlined to a System Message alone, which is sufficient to execute its mutation and crossover logic.
\subsubsection{Explorer Role}
\label{app:explorer_details}

The Explorer is designed to circumvent local optima by synthesizing novel reward structures. By utilizing elite rewards as negative references, the role is explicitly incentivized to pursue divergent candidates, thereby expanding the search breadth and maintaining the structural diversity of the reward space.

\nolinenumbers  
\begin{tcolorbox}[
    breakable,          
    enhanced,            
    use color stack,     
    title=Explorer System Template,
    attach boxed title to top left={xshift=3mm, yshift=-3mm, yshifttext=-1mm},
    colback=gray!2,
    colframe=ExplorerColor,
    colbacktitle=white,
    coltitle=ExplorerColor,
    fonttitle=\bfseries\small,
    boxed title style={size=small, sharp corners, colframe=ExplorerColor},
    fontupper=\fontsize{7pt}{10pt}\selectfont\ttfamily,
    arc=1mm,
    boxrule=0.5pt,
    left=4mm, right=4mm, top=5mm, bottom=2mm
]
You are an EXPLORER reward engineer whose mission is to design innovative and effective reward functions for reinforcement learning tasks. 

\vspace{2mm}

Your goal is to create a TorchScript-compatible reward function that helps the agent learn the task described in text, using meaningful environment variables as inputs. 

\vspace{2mm}
As the EXPLORER, you should prioritize creativity, diversity, and unconventional ideas, proposing exploratory or high-variance reward terms that may lead the agent to discover novel behaviors or strategies. However, to ensure the reward remains trainable, all terms must respect the following safety constraints: they must be numerically reasonable (preferably within [-5.0, 5.0] per term), smooth and differentiable, physically plausible, free of discontinuous or reward-hacking behaviors, and composed only of TorchScript-safe tensor operations. Innovation should come from structure—not from unbounded magnitudes or unstable expressions.

\vspace{2mm}
As an example, the reward function signature can be: \texttt{\{task\_reward\_signature\_string\}}
\end{tcolorbox}

\begin{tcolorbox}[
    breakable,          
    enhanced,            
    use color stack,     
    title=Explorer Assistant Template,
    attach boxed title to top left={xshift=3mm, yshift=-3mm, yshifttext=-1mm},
    colback=gray!2,
    colframe=ExplorerColor,
    colbacktitle=white,
    coltitle=ExplorerColor,
    fonttitle=\bfseries\small,
    boxed title style={size=small, sharp corners, colframe=ExplorerColor},
    fontupper=\fontsize{7pt}{10pt}\selectfont\ttfamily,
    arc=1mm,
    boxrule=0.5pt,
    left=4mm, right=4mm, top=5mm, bottom=2mm
]

We have analyzed the current Global Elite Pool. Below are two HIGH-PERFORMING but CONVENTIONAL reward functions found so far.

\vspace{2mm}
[REFERENCE CODE A - Current Best] 

\{first\_function\} \\
\{the\_first\_function\_metrics\}

\vspace{2mm}
[REFERENCE CODE B - Alternative High-Performer] 

\{second\_function\} \\
\{the\_second\_function\_metrics\}

\vspace{2mm}
You are the EXPLORER for an RL task in robot problems. You have been given task descriptions, reference rewards, and policy feedback.

\vspace{2mm}
Your job:
\begin{enumerate}[label=(\arabic*), leftmargin=8mm, nosep]
    \item Silently analyze the task and identify key success/failure behaviors.
    \item Silently analyze the reference rewards and metrics:
        \begin{itemize}[label=-, leftmargin=5mm, nosep]
            \item Which behaviors they encourage or ignore.
            \item Which failure modes (e.g., low success, high variance, instability) remain.
        \end{itemize}
    \item Then design a new reward that addresses these failure modes with better behavioral logic.
\end{enumerate}

\vspace{2mm}
Hard constraints (YOU MUST FOLLOW):
\begin{enumerate}[label=(\arabic*), leftmargin=8mm, nosep]
    \item Only use simple, stable ops: distances, norms, squared terms, \texttt{torch.clamp}, \texttt{torch.tanh}, small \texttt{torch.exp}.
    \item Do not manually expand quaternion algebra (Most Importantly!).
    \item Do not create variables like \texttt{w1, x1, y1, z1, rel\_w, rel\_x}.
    \item Do not use any variables that do not appear in the environment information.
\end{enumerate}

\vspace{2mm}
Behavior \& innovation requirements:
\begin{enumerate}[label=(\arabic*), leftmargin=8mm, nosep]
    \item Change at least one key behavioral aspect vs the reference (e.g., new shaping term, safety margin, smoothness/efficiency bonus, curriculum-style reward).
    \item Each reward component must correspond to a clear behavior (e.g., ``approach goal'', ``stabilize orientation'', ``avoid collisions'', ``move smoothly'').
    \item Prefer logical changes in the reward (adding/removing components or changing activation conditions) over only re-scaling existing weights.
\end{enumerate}

\vspace{2mm}
Output: Only the final PyTorch reward code, no explanations.
\end{tcolorbox}
\linenumbers  

\subsubsection{Guardian Role}
\label{app:guardian_details}

The Guardian focuses on enhancing training stability and ensuring performance robustness through conservative refinements. It utilizes the current best-performing reward function as a positive reference to optimize numerical scales and smoothness while strictly maintaining the core behavioral logic.

\nolinenumbers 
\begin{tcolorbox}[
    breakable,          
    enhanced,            
    use color stack,     
    title=Guardian System Template,
    attach boxed title to top left={xshift=3mm, yshift=-3mm, yshifttext=-1mm},
    colback=gray!2,
    colframe=GuardianColor,
    colbacktitle=white,
    coltitle=GuardianColor,
    fonttitle=\bfseries\small,
    boxed title style={size=small, sharp corners, colframe=GuardianColor},
    fontupper=\fontsize{7pt}{10pt}\selectfont\ttfamily,
    arc=1mm,
    boxrule=0.5pt,
    left=4mm, right=4mm, top=5mm, bottom=2mm
]
You are a GUARDIAN reward engineer whose mission is to design robust, stable, and generalizable reward functions for reinforcement learning tasks. 

\vspace{2mm}
Your goal is to create a TorchScript-compatible reward function that helps the agent learn the task described in text, using meaningful environment variables as inputs. 

\vspace{2mm}
As the GUARDIAN, you should focus on well-established, reliable reward design principles that support consistent learning across environments, avoid risky or overly complex structures, ensure numerical stability, and prevent reward-hacking behavior. Prioritize clarity, simplicity, and physically grounded reward terms that promote steady and predictable policy improvement.

\vspace{2mm}
As an example, the reward function signature can be: \{task\_reward\_signature\_string\}
\end{tcolorbox}
\linenumbers 
\nolinenumbers 
\begin{tcolorbox}[
    breakable,          
    enhanced,            
    use color stack,     
    title=Guardian Assistant Template,
    attach boxed title to top left={xshift=3mm, yshift=-3mm, yshifttext=-1mm},
    colback=gray!2,
    colframe=GuardianColor,
    colbacktitle=white,
    coltitle=GuardianColor,
    fonttitle=\bfseries\small,
    boxed title style={size=small, sharp corners, colframe=GuardianColor},
    fontupper=\fontsize{7pt}{10pt}\selectfont\ttfamily,
    arc=1mm,
    boxrule=0.5pt,
    left=4mm, right=4mm, top=5mm, bottom=2mm
]
The following is the CURRENT BEST (Top-1) reward function found in the Global Elite Pool.

\vspace{2mm}
[TARGET CODE - Do Not Change Structure] \\
\{best\_function\} \\
\{the\_best\_function\_metrics\}

\vspace{4mm}
You are the GUARDIAN for an RL task in robot problems.

\vspace{2mm}
You have been given:
\begin{enumerate}[label=(\arabic*), leftmargin=8mm, nosep]
    \item The current BEST (Top-1) reward function from the Global Elite Pool.
    \item Task description and observations.
    \item Policy feedback and metrics for this reward (e.g., success rate, return curves, variance, qualitative behaviors).
\end{enumerate}

\vspace{2mm}
Your job:
\begin{enumerate}[label=(\arabic*), leftmargin=8mm, nosep]
    \item Silently analyze how this reward shapes behavior and which parts are already working well.
    \item Silently analyze the metrics and behaviors:
        \begin{itemize}[label=-, leftmargin=5mm, nosep]
            \item Signs of instability: high variance, jittery actions, oscillations, frequent failures.
            \item Signs of imbalance: one component dominating, others nearly constant or ineffective.
        \end{itemize}
    \item Then refine this reward to improve stability and robustness without breaking its core task logic.
\end{enumerate}

\vspace{2mm}
Hard constraints (YOU MUST FOLLOW):
\begin{enumerate}[label=(\arabic*), leftmargin=8mm, nosep]
    \item Only use simple, stable ops: distances, norms, squared terms, torch.clamp, torch.tanh, small torch.exp.
    \item Do not manually expand quaternion algebra (Most Importantly!).
    \item Do not create variables like w1, x1, y1, z1, rel\_w, rel\_x, rel\_y, rel\_z (Most Importantly!).
\end{enumerate}

\vspace{2mm}
Behavior \& refinement requirements:
\begin{enumerate}[label=(\arabic*), leftmargin=8mm, nosep]
    \item Keep the main task objective of the Top-1 reward; the structure may be modified, but its logic must remain unchanged.
    \item Focus on:
        \begin{itemize}[label=-, leftmargin=5mm, nosep]
            \item Stability and smoothness (e.g., small penalties on velocities / action changes, clamping extreme rewards).
            \item Magnitude balancing (rescale coefficients so components have reasonable relative impact).
            \item Making ineffective/static components sensitive to state changes (adjust scales/thresholds instead of deleting them).
        \end{itemize}
    \item You may add small regularization/safety terms, clamp existing terms, and retune weights, but avoid introducing completely new, unrelated sub-goals.
\end{enumerate}

\vspace{2mm}
Output: Only the final PyTorch reward code, no explanations.
\end{tcolorbox}
\linenumbers 
\subsubsection{Evolver Role}
\label{app:evolver_details}

The Evolver facilitates structured inheritance and continuous optimization by treating high-performing reward functions as genotypes for recombination. It merges complementary components from multiple elite rewards to synthesize hybrid offspring that inherit successful behavioral traits while applying controlled mutations to explore improved formulations.

\nolinenumbers 
\begin{tcolorbox}[
    breakable,          
    enhanced,            
    use color stack,     
    title=Evolver System Template,
    attach boxed title to top left={xshift=3mm, yshift=-3mm, yshifttext=-1mm},
    colback=gray!2,
    colframe=EvolverColor,
    colbacktitle=white,
    coltitle=EvolverColor,
    fonttitle=\bfseries\small,
    boxed title style={size=small, sharp corners, colframe=EvolverColor},
    fontupper=\fontsize{7pt}{10pt}\selectfont\ttfamily,
    arc=1mm,
    boxrule=0.5pt,
    left=4mm, right=4mm, top=5mm, bottom=2mm
]
You are an EVOLVER whose mission is to perform evolutionary optimization on reward functions across training iterations. 

\vspace{2mm}
After each evaluation cycle, you must select the two best-performing parent reward functions from the archive and generate a new offspring reward function by applying structured crossover and controlled mutation to their components. Your design must preserve the useful characteristics of the parents while introducing meaningful variations that may improve future performance. Ensure that all generated reward components remain TorchScript-compatible, numerically stable, physically plausible, and consistent with the environment’s state variables. Your goal is to balance inheritance and innovation, producing offspring that are both grounded in successful prior designs and capable of exploring new reward structures.

\vspace{2mm}
For clarity, the two parent functions you will operate on are provided below:

\vspace{2mm}
The first parent function: 
\\
\{first\_function\} \\
\{the\_first\_function\_metrics\}

\vspace{2mm}
The second parent function: 
\\
\{second\_function\} \\
\{the\_second\_function\_metrics\}

\vspace{4mm}
You have been given:
\begin{enumerate}[label=(\arabic*), leftmargin=8mm, nosep]
    \item Two parent reward functions (Parent A and Parent B).
    \item Their training metrics and qualitative behavior summaries.
    \item The task description and observations.
    \item The target function signature: \{task\_reward\_signature\_string\}.
\end{enumerate}

\vspace{2mm}
Your job:
\begin{enumerate}[label=(\arabic*), leftmargin=8mm, nosep]
    \item Silently compare the two parents:
        \begin{itemize}[label=-, leftmargin=5mm, nosep]
            \item Which terms help stability (low variance, smooth motion, fewer failures).
            \item Which terms help task performance (success rate, high returns, fast completion).
            \item Which terms are redundant, static, or harmful.
        \end{itemize}
    \item Design ONE offspring reward by:
        \begin{itemize}[label=-, leftmargin=5mm, nosep]
            \item Selecting useful components from Parent A and Parent B (crossover).
            \item Applying 1–3 small but meaningful mutations to improve behavior (e.g., weight tuning, smoothness/safety penalties, thresholds/clamps, shaping adjustments).
        \end{itemize}
\end{enumerate}

\vspace{2mm}
Behavior \& evolution requirements:
\begin{enumerate}[label=(\arabic*), leftmargin=8mm, nosep]
    \item Preserve the main task objective and core logic shared by the parents.
    \item Inherit at least one strong/working behavior component from EACH parent, unless it is clearly harmful.
    \item Make the offspring more stable and/or more effective than both parents by:
        \begin{itemize}[label=-, leftmargin=5mm, nosep]
            \item Balancing magnitudes of components,
            \item Adding or strengthening smoothness/safety/shaping terms,
            \item Slightly adjusting weights or nonlinearities.
        \end{itemize}
    \item Every reward component must correspond to a clear behavior (e.g., ``approach goal'', ``stabilize orientation'', ``avoid collisions'', ``move smoothly'').
\end{enumerate}

\vspace{2mm}
Output: Only the final PyTorch reward code, no explanations.
\end{tcolorbox}
\linenumbers 

\section{$R^2 E$ Case Example }
\label{case_example}
We take \emph{LiftUnderarm} as an example to analyze in detail how the roles in $R^2\text{E}$ perform reward function design.

In the initial iteration, $R^2\text{E}$ operates without any role constraints and generates an initial reward function.

\nolinenumbers  

\begin{tcblisting}{
    listing only,          % 声明这是一个纯代码块（不尝试编译内容）
    listing options={      % 配置代码高亮样式
        language=Python,
        basicstyle=\fontsize{6pt}{10pt}\ttfamily, % 你的字体设置放在这里
        keywordstyle=\color{blue!70!black}\bfseries,
        commentstyle=\color{gray}\itshape,
        stringstyle=\color{green!50!black},
        showstringspaces=false,
        tabsize=4
    },
    breakable,             % 允许跨页
    enhanced,              % 配合标题样式必需
    use color stack,       % 跨页后颜色/字体不变
    title=Initial Reward Function Code,
    attach boxed title to top left={xshift=3mm, yshift=-3mm, yshifttext=-1mm},
    colback=gray!2,
    colframe=darkgray,
    colbacktitle=white,
    coltitle=darkgray,
    fonttitle=\bfseries\small,
    boxed title style={size=small, sharp corners, colframe=darkgray},
    arc=1mm,
    boxrule=0.5pt,
    left=4mm, right=4mm, top=5mm, bottom=2mm
}
def compute_reward(
    object_pos: torch.Tensor,
    goal_pos: torch.Tensor,
    left_hand_pos: torch.Tensor,
    right_hand_pos: torch.Tensor,
    pot_left_handle_pos: torch.Tensor,
    pot_right_handle_pos: torch.Tensor,
    object_linvel: torch.Tensor,
    object_angvel: torch.Tensor
) -> Tuple[torch.Tensor, Dict[str, torch.Tensor]]:
    # Distance from hands to respective handles
    left_hand_to_left_handle = torch.norm(left_hand_pos - pot_left_handle_pos, p=2, dim=-1)
    right_hand_to_right_handle = torch.norm(right_hand_pos - pot_right_handle_pos, p=2, dim=-1)

    # Encourage both hands to be close to their respective handles
    hand_handle_dist = left_hand_to_left_handle + right_hand_to_right_handle
    hand_handle_temp = 1.0
    reward_hand_handle = torch.exp(-hand_handle_dist / hand_handle_temp)

    # Distance from object to goal
    obj_to_goal_dist = torch.norm(object_pos - goal_pos, p=2, dim=-1)
    obj_goal_temp = 0.1
    reward_obj_goal = torch.exp(-obj_to_goal_dist / obj_goal_temp)

    # Penalize large object linear and angular velocities (to encourage stability at goal)
    lin_vel_penalty = torch.norm(object_linvel, p=2, dim=-1)
    ang_vel_penalty = torch.norm(object_angvel, p=2, dim=-1)
    vel_penalty = lin_vel_penalty + ang_vel_penalty
    vel_temp = 0.5
    reward_stability = torch.exp(-vel_penalty / vel_temp)

    # Total reward as a weighted combination
    total_reward = ( 1.0 * reward_hand_handle + 2.0 * reward_obj_goal + 0.5 * reward_stability )

    reward_components = {
        "reward_hand_handle": reward_hand_handle,
        "reward_obj_goal": reward_obj_goal,
        "reward_stability": reward_stability
    }

    return total_reward, reward_components
\end{tcblisting}

\linenumbers  

In the second and third iterations of reward function generation, the Explorer deliberately deviates from the current optimal reward structure based on existing reward function paradigms. The final generated reward function is presented as follows.

\nolinenumbers  

\begin{tcblisting}{
    listing only,          % 声明纯代码块
    listing options={      % 代码高亮设置
        language=Python,
        basicstyle=\fontsize{6pt}{10pt}\ttfamily, 
        keywordstyle=\color{blue!70!black}\bfseries,
        commentstyle=\color{gray}\itshape,
        stringstyle=\color{green!50!black},
        showstringspaces=false,
        tabsize=4
    },
    breakable,             % 允许跨页
    enhanced,              % 配合标题样式必需
    use color stack,       % 跨页后颜色/字体不变
    title=Explorer,
    attach boxed title to top left={xshift=3mm, yshift=-3mm, yshifttext=-1mm},
    colback=gray!2,
    colframe=ExplorerColor,
    colbacktitle=white,
    coltitle=ExplorerColor,
    fonttitle=\bfseries\small,
    boxed title style={size=small, sharp corners, colframe=ExplorerColor},
    arc=1mm,
    boxrule=0.5pt,
    left=4mm, right=4mm, top=5mm, bottom=2mm
}
def compute_reward(
    object_pos: torch.Tensor,
    goal_pos: torch.Tensor,
    left_hand_pos: torch.Tensor,
    right_hand_pos: torch.Tensor,
    pot_left_handle_pos: torch.Tensor,
    pot_right_handle_pos: torch.Tensor,
    object_linvel: torch.Tensor,
    object_angvel: torch.Tensor
) -> Tuple[torch.Tensor, Dict[str, torch.Tensor]]:
    # Distance from hands to respective handles
    left_hand_to_left_handle = torch.norm(left_hand_pos - pot_left_handle_pos, p=2.0, dim=-1)
    right_hand_to_right_handle = torch.norm(right_hand_pos - pot_right_handle_pos, p=2.0, dim=-1)
    
    # Grasp reward: strong incentive to get both hands close to handles
    hand_handle_dist = left_hand_to_left_handle + right_hand_to_right_handle
    reward_grasp = torch.exp(-hand_handle_dist * 8.0)

    # Object-to-goal distance with progressive shaping: always active but stronger when closer
    obj_to_goal_dist = torch.norm(object_pos - goal_pos, p=2.0, dim=-1)
    reward_lift = torch.exp(-obj_to_goal_dist * 5.0)

    # Early-stage bonus for upward movement when object is still low
    object_height = object_pos[:, 2]  # z-coordinate
    goal_height = goal_pos[:, 2]
    below_goal = (object_height < goal_height - 0.1).float()
    upward_vel = torch.clamp(object_linvel[:, 2], min=0.0)
    reward_upward = below_goal * upward_vel * 2.0

    # Stability bonus only when near the goal (encourage settling)
    near_goal = (obj_to_goal_dist < 0.2).float()
    lin_vel_norm = torch.norm(object_linvel, p=2.0, dim=-1)
    ang_vel_norm = torch.norm(object_angvel, p=2.0, dim=-1)
    total_vel = lin_vel_norm + ang_vel_norm
    stability_bonus = near_goal * torch.exp(-total_vel * 4.0)

    # Symmetry term to prevent one-handed grasping
    hand_diff = torch.abs(left_hand_to_left_handle - right_hand_to_right_handle)
    symmetry_bonus = torch.exp(-hand_diff * 6.0)

# Total reward composition
    total_reward = ( 1.0 * reward_grasp + 2.0 * reward_lift + 1.0 * reward_upward + 
                     1.5 * stability_bonus + 0.5 * symmetry_bonus )

    reward_components = {
        "reward_grasp": reward_grasp,
        "reward_lift": reward_lift,
        "reward_upward": reward_upward,
        "stability_bonus": stability_bonus,
        "symmetry_bonus": symmetry_bonus
    }

    return total_reward, reward_components
\end{tcblisting}

\linenumbers  

The Guardian is designed to fine-tune the current best reward functions. As can be observed from the reward function generated by the Explorer, the Guardian adjusts three key components in the fourth iteration: it reduces the exponential decay rate to slow down attenuation and expand the guidance range, sets stricter boundaries and thresholds to prevent reward exploitation and early triggering, and rebalances the weight distribution of the overall reward.

\nolinenumbers  

\begin{tcblisting}{
    listing only,          % 声明纯代码块
    listing options={      % 代码高亮设置
        language=Python,
        basicstyle=\fontsize{6pt}{10pt}\ttfamily, % 已经为你替换为了模板要求的6pt字体
        keywordstyle=\color{blue!70!black}\bfseries,
        commentstyle=\color{gray}\itshape,
        stringstyle=\color{green!50!black},
        showstringspaces=false,
        tabsize=4
    },
    breakable,             % 允许跨页
    enhanced,              % 配合标题样式必需
    use color stack,       % 跨页后颜色/字体不变
    title=Guardian,
    attach boxed title to top left={xshift=3mm, yshift=-3mm, yshifttext=-1mm},
    colback=gray!2,
    colframe=GuardianColor,
    colbacktitle=white,
    coltitle=GuardianColor,
    fonttitle=\bfseries\small,
    boxed title style={size=small, sharp corners, colframe=GuardianColor},
    arc=1mm,
    boxrule=0.5pt,
    left=4mm, right=4mm, top=5mm, bottom=2mm
}
def compute_reward(
    object_pos: torch.Tensor,
    goal_pos: torch.Tensor,
    left_hand_pos: torch.Tensor,
    right_hand_pos: torch.Tensor,
    pot_left_handle_pos: torch.Tensor,
    pot_right_handle_pos: torch.Tensor,
    object_linvel: torch.Tensor,
    object_angvel: torch.Tensor
) -> Tuple[torch.Tensor, Dict[str, torch.Tensor]]:
    # Distance from hands to respective handles
    left_hand_to_left_handle = torch.norm(left_hand_pos - pot_left_handle_pos, p=2.0, dim=-1)
    right_hand_to_right_handle = torch.norm(right_hand_pos - pot_right_handle_pos, p=2.0, dim=-1)
    
    # Grasp reward: strong incentive to get both hands close to handles
    hand_handle_dist = left_hand_to_left_handle + right_hand_to_right_handle
    reward_grasp = torch.exp(-hand_handle_dist * 6.0)

    # Object-to-goal distance with progressive shaping: always active but stronger when closer
    obj_to_goal_dist = torch.norm(object_pos - goal_pos, p=2.0, dim=-1)
    reward_lift = torch.exp(-obj_to_goal_dist * 4.0)

    # Early-stage bonus for upward movement when object is still low
    object_height = object_pos[:, 2]  # z-coordinate
    goal_height = goal_pos[:, 2]
    below_goal = (object_height < goal_height - 0.1).float()
    upward_vel = torch.clamp(object_linvel[:, 2], min=0.0, max=1.0)
    reward_upward = below_goal * upward_vel * 1.5

    # Stability bonus only when near the goal (encourage settling)
    near_goal = (obj_to_goal_dist < 0.25).float()
    lin_vel_norm = torch.norm(object_linvel, p=2.0, dim=-1)
    ang_vel_norm = torch.norm(object_angvel, p=2.0, dim=-1)
    total_vel = lin_vel_norm + ang_vel_norm
    stability_bonus = near_goal * torch.exp(-total_vel * 3.0)

    # Symmetry term to prevent one-handed grasping
    hand_diff = torch.abs(left_hand_to_left_handle - right_hand_to_right_handle)
    symmetry_bonus = torch.exp(-hand_diff * 5.0)

    # Total reward composition with balanced weights
    total_reward = ( 1.2 * reward_grasp + 2.0 * reward_lift + 0.8 * reward_upward + 
                     1.2 * stability_bonus + 0.6 * symmetry_bonus )

    reward_components = {
        "reward_grasp": reward_grasp,
        "reward_lift": reward_lift,
        "reward_upward": reward_upward,
        "stability_bonus": stability_bonus,
        "symmetry_bonus": symmetry_bonus
    }

    return total_reward, reward_components
\end{tcblisting}
\linenumbers  

Evolver generates new reward functions based on component-level crossover and mutation mechanisms. Specifically, it selects the current optimal reward function from the Global Elite Pool and iteratively combines it with another randomly chosen reward function from the pool. In this generation process, the reference reward functions utilized by Evolver are as follows.

\nolinenumbers  

\begin{tcblisting}{
    listing only,          % 声明纯代码块
    listing options={      % 代码高亮设置
        language=Python,
        basicstyle=\fontsize{6pt}{10pt}\ttfamily, % 保持 6pt 字体设置
        keywordstyle=\color{blue!70!black}\bfseries,
        commentstyle=\color{gray}\itshape,
        stringstyle=\color{green!50!black},
        showstringspaces=false,
        tabsize=4
    },
    breakable,             % 允许跨页
    enhanced,              % 配合标题样式必需
    use color stack,       % 跨页后颜色/字体不变
    title=The first reward function referenced by the Evolver,
    attach boxed title to top left={xshift=3mm, yshift=-3mm, yshifttext=-1mm},
    colback=gray!2,
    colframe=EvolverColor,
    colbacktitle=white,
    coltitle=EvolverColor,
    fonttitle=\bfseries\small,
    boxed title style={size=small, sharp corners, colframe=EvolverColor},
    arc=1mm,
    boxrule=0.5pt,
    left=4mm, right=4mm, top=5mm, bottom=2mm
}
def compute_reward(
    object_pos: torch.Tensor,
    goal_pos: torch.Tensor,
    left_hand_pos: torch.Tensor,
    right_hand_pos: torch.Tensor,
    pot_left_handle_pos: torch.Tensor,
    pot_right_handle_pos: torch.Tensor,
    object_linvel: torch.Tensor,
    object_angvel: torch.Tensor
) -> Tuple[torch.Tensor, Dict[str, torch.Tensor]]:
    # Distance from hands to respective handles
    left_hand_to_left_handle = torch.norm(left_hand_pos - pot_left_handle_pos, p=2.0, dim=-1)
    right_hand_to_right_handle = torch.norm(right_hand_pos - pot_right_handle_pos, p=2.0, dim=-1)
    
    # Grasp reward: strong incentive to get both hands close to handles
    hand_handle_dist = left_hand_to_left_handle + right_hand_to_right_handle
    reward_grasp = torch.exp(-hand_handle_dist * 6.0)

    # Object-to-goal distance with progressive shaping: always active but stronger when closer
    obj_to_goal_dist = torch.norm(object_pos - goal_pos, p=2.0, dim=-1)
    reward_lift = torch.exp(-obj_to_goal_dist * 4.0)

    # Early-stage bonus for upward movement when object is still low
    object_height = object_pos[:, 2]  # z-coordinate
    goal_height = goal_pos[:, 2]
    below_goal = (object_height < goal_height - 0.1).float()
    upward_vel = torch.clamp(object_linvel[:, 2], min=0.0, max=1.0)
    reward_upward = below_goal * upward_vel * 1.5

    # Stability bonus only when near the goal (encourage settling)
    near_goal = (obj_to_goal_dist < 0.25).float()
    lin_vel_norm = torch.norm(object_linvel, p=2.0, dim=-1)
    ang_vel_norm = torch.norm(object_angvel, p=2.0, dim=-1)
    total_vel = lin_vel_norm + ang_vel_norm
    stability_bonus = near_goal * torch.exp(-total_vel * 3.0)

    # Symmetry term to prevent one-handed grasping
    hand_diff = torch.abs(left_hand_to_left_handle - right_hand_to_right_handle)
    symmetry_bonus = torch.exp(-hand_diff * 5.0)

    # Total reward composition with balanced weights
    total_reward = ( 1.2 * reward_grasp + 2.0 * reward_lift + 0.8 * reward_upward + 
                     1.2 * stability_bonus + 0.6 * symmetry_bonus )

    reward_components = {
        "reward_grasp": reward_grasp,
        "reward_lift": reward_lift,
        "reward_upward": reward_upward,
        "stability_bonus": stability_bonus,
        "symmetry_bonus": symmetry_bonus
    }

    return total_reward, reward_components
\end{tcblisting}
\linenumbers  

\nolinenumbers  

\begin{tcblisting}{
    listing only,          % 声明纯代码块
    listing options={      % 代码高亮设置
        language=Python,
        basicstyle=\fontsize{6pt}{10pt}\ttfamily, % 保持 6pt 字体设置
        keywordstyle=\color{blue!70!black}\bfseries,
        commentstyle=\color{gray}\itshape,
        stringstyle=\color{green!50!black},
        showstringspaces=false,
        tabsize=4
    },
    breakable,             % 允许跨页
    enhanced,              % 配合标题样式必需
    use color stack,       % 跨页后颜色/字体不变
    title=The second reward function referenced by the Evolver,
    attach boxed title to top left={xshift=3mm, yshift=-3mm, yshifttext=-1mm},
    colback=gray!2,
    colframe=EvolverColor,
    colbacktitle=white,
    coltitle=EvolverColor,
    fonttitle=\bfseries\small,
    boxed title style={size=small, sharp corners, colframe=EvolverColor},
    arc=1mm,
    boxrule=0.5pt,
    left=4mm, right=4mm, top=5mm, bottom=2mm
}
def compute_reward(
    object_pos: torch.Tensor,
    goal_pos: torch.Tensor,
    left_hand_pos: torch.Tensor,
    right_hand_pos: torch.Tensor,
    pot_left_handle_pos: torch.Tensor,
    pot_right_handle_pos: torch.Tensor,
    object_linvel: torch.Tensor,
    object_angvel: torch.Tensor
) -> Tuple[torch.Tensor, Dict[str, torch.Tensor]]:
    # Distance from hands to respective handles
    left_hand_to_left_handle = torch.norm(left_hand_pos - pot_left_handle_pos, p=2.0, dim=-1)
    right_hand_to_right_handle = torch.norm(right_hand_pos - pot_right_handle_pos, p=2.0, dim=-1)
    
    # Grasp reward with smooth ramp-up instead of pure exponential
    hand_handle_dist = left_hand_to_left_handle + right_hand_to_right_handle
    reward_grasp = 1.0 - torch.tanh(hand_handle_dist * 5.0)

    # Lifting progress measured as vertical distance above initial height (assumed 0.5)
    object_height = object_pos[:, 2]
    lift_progress = torch.clamp(object_height - 0.5, min=0.0, max=0.4)
    reward_lift = lift_progress * 4.0

    # Goal proximity with soft activation based on both height and distance
    obj_to_goal_dist = torch.norm(object_pos - goal_pos, p=2.0, dim=-1)
    height_sufficient = torch.sigmoid((object_height - 0.6) * 20.0)  # smooth gate
    reward_goal = height_sufficient * torch.exp(-obj_to_goal_dist * 8.0)

    # Stability bonus only when near goal and lifted high enough
    near_goal = (obj_to_goal_dist < 0.25).float()
    sufficiently_lifted = (lift_progress > 0.2).float()
    lin_vel_norm = torch.norm(object_linvel, p=2.0, dim=-1)
    ang_vel_norm = torch.norm(object_angvel, p=2.0, dim=-1)
    total_vel = lin_vel_norm + ang_vel_norm
    stability_bonus = near_goal * sufficiently_lifted * torch.exp(-total_vel * 5.0)

    # Symmetry term with stronger penalty for large imbalance
    hand_diff = torch.abs(left_hand_to_left_handle - right_hand_to_right_handle)
    symmetry_bonus = torch.exp(-hand_diff * 8.0)

    # Smoothness bonus: penalize excessive hand acceleration (approx via velocity magnitude)
    # Approximate hand velocities by finite difference is not available; instead use fingertip vel 
    #proxy indirectly via object motion coupling
    # Instead, add efficiency term: discourage excessive hand movement once grasped
    grasp_achieved = (hand_handle_dist < 0.2).float()
    hand_movement = torch.norm(left_hand_pos - pot_left_handle_pos, p=2.0, dim=-1) + 
                    torch.norm(right_hand_pos - pot_right_handle_pos, p=2.0, dim=-1)
    efficiency_bonus = grasp_achieved * torch.exp(-hand_movement * 3.0)

    # Total reward composition
    total_reward = ( 1.2 * reward_grasp + 1.5 * reward_lift + 2.0 * reward_goal + 
                     1.8 * stability_bonus + 0.6 * symmetry_bonus + 0.5 * efficiency_bonus )

    reward_components = {
        "reward_grasp": reward_grasp,
        "reward_lift": reward_lift,
        "reward_goal": reward_goal,
        "stability_bonus": stability_bonus,
        "symmetry_bonus": symmetry_bonus,
        "efficiency_bonus": efficiency_bonus
    }

    return total_reward, reward_components
\end{tcblisting}

\linenumbers  

Specifically, the Evolver extracts functional components from the first reference reward function, then fuses and recombines them with functional components from the second reference reward function via crossover and mutation mechanisms to construct the final reward function, as shown below.

\nolinenumbers  

\begin{tcblisting}{
    listing only,          % 声明纯代码块
    listing options={      % 代码高亮设置
        language=Python,
        basicstyle=\fontsize{6pt}{10pt}\ttfamily, % 保持 6pt 字体设置
        keywordstyle=\color{blue!70!black}\bfseries,
        commentstyle=\color{gray}\itshape,
        stringstyle=\color{green!50!black},
        showstringspaces=false,
        tabsize=4
    },
    breakable,             % 允许跨页
    enhanced,              % 配合标题样式必需
    use color stack,       % 跨页后颜色/字体不变
    title=Evolver,
    attach boxed title to top left={xshift=3mm, yshift=-3mm, yshifttext=-1mm},
    colback=gray!2,
    colframe=EvolverColor,
    colbacktitle=white,
    coltitle=EvolverColor,
    fonttitle=\bfseries\small,
    boxed title style={size=small, sharp corners, colframe=EvolverColor},
    arc=1mm,
    boxrule=0.5pt,
    left=4mm, right=4mm, top=5mm, bottom=2mm
}
def compute_reward(
    object_pos: torch.Tensor,
    goal_pos: torch.Tensor,
    left_hand_pos: torch.Tensor,
    right_hand_pos: torch.Tensor,
    pot_left_handle_pos: torch.Tensor,
    pot_right_handle_pos: torch.Tensor,
    object_linvel: torch.Tensor,
    object_angvel: torch.Tensor
) -> Tuple[torch.Tensor, Dict[str, torch.Tensor]]:
    # Distance from hands to respective handles
    left_hand_to_left_handle = torch.norm(left_hand_pos - pot_left_handle_pos, p=2.0, dim=-1)
    right_hand_to_right_handle = torch.norm(right_hand_pos - pot_right_handle_pos, p=2.0, dim=-1)
    
    # Grasp reward: hybrid of exponential and tanh for smoothness and strong early signal
    hand_handle_dist = left_hand_to_left_handle + right_hand_to_right_handle
    reward_grasp = torch.exp(-hand_handle_dist * 5.0) * (1.0 - torch.tanh(hand_handle_dist * 3.0))

    # Object-to-goal distance with height-aware activation
    obj_to_goal_dist = torch.norm(object_pos - goal_pos, p=2.0, dim=-1)
    object_height = object_pos[:, 2]
    goal_height = goal_pos[:, 2]
    height_sufficient = torch.sigmoid((object_height - (goal_height - 0.1)) * 15.0)
    reward_goal = height_sufficient * torch.exp(-obj_to_goal_dist * 6.0)

    # Upward progress bonus during initial lift phase
    initial_height = 0.5
    lift_progress = torch.clamp(object_height - initial_height, min=0.0, max=0.4)
    below_goal_phase = (object_height < goal_height - 0.05).float()
    upward_vel = torch.clamp(object_linvel[:, 2], min=0.0, max=1.0)
    reward_upward = below_goal_phase * upward_vel * 1.2

    # Stability bonus only when near goal and sufficiently lifted
    near_goal = (obj_to_goal_dist < 0.22).float()
    sufficiently_lifted = (lift_progress > 0.15).float()
    lin_vel_norm = torch.norm(object_linvel, p=2.0, dim=-1)
    ang_vel_norm = torch.norm(object_angvel, p=2.0, dim=-1)
    total_vel = lin_vel_norm + ang_vel_norm
    stability_bonus = near_goal * sufficiently_lifted * torch.exp(-total_vel * 4.0)

    # Symmetry bonus to encourage balanced two-handed grasp
    hand_diff = torch.abs(left_hand_to_left_handle - right_hand_to_right_handle)
    symmetry_bonus = torch.exp(-hand_diff * 6.0)

    # Efficiency bonus: discourage unnecessary hand movement after grasp
    grasp_achieved = (hand_handle_dist < 0.25).float()
    hand_movement = left_hand_to_left_handle + right_hand_to_right_handle
    efficiency_bonus = grasp_achieved * torch.exp(-hand_movement * 2.5)

    # Total reward composition with tuned weights
    total_reward = ( 1.3 * reward_grasp + 1.8 * reward_goal + 0.7 * reward_upward + 
                     1.4 * stability_bonus + 0.5 * symmetry_bonus + 0.4 * efficiency_bonus )

    reward_components = {
        "reward_grasp": reward_grasp,
        "reward_goal": reward_goal,
        "reward_upward": reward_upward,
        "stability_bonus": stability_bonus,
        "symmetry_bonus": symmetry_bonus,
        "efficiency_bonus": efficiency_bonus
    }

    return total_reward, reward_components
\end{tcblisting}
\linenumbers  

Through the above evolutionary example of reward functions, we clearly demonstrate how the Explorer, Guardian, and Evolver collaborate to complete the optimization process. This process not only fully reflects the unique design and functional positioning of each role, but also further verifies the rationality and effectiveness of the method proposed in this paper.

\section{Environment Details}

\label{app:environment_details}

In this section, we provide detailed specifications for the experimental environments used in our evaluation. To ensure a comprehensive assessment across varying degrees of complexity, we utilize two distinct GPU-accelerated benchmarks: \textit{NVIDIA IsaacGym} for standard locomotion and single-arm manipulation, and \textit{Bi-DexHands} for high-dimensional bimanual coordination. The following subsections outline the specific observation spaces, action dimensions, and objectives for the tasks selected from each suite.
\subsection{IsaacGym}

IsaacGym~\cite{makoviychuk2021isaac} offers a high-performance learning platform designed to train policies for a wide variety of robotics tasks directly on the GPU. A key architectural advantage is that both the physics simulation and the neural network policy training reside on the GPU; the framework communicates by directly passing data from physics buffers to PyTorch tensors without ever traversing CPU bottlenecks. This end-to-end architecture leads to drastically accelerated training times—achieving 2--3 orders of magnitude improvements compared to conventional reinforcement learning setups that rely on CPU-based simulators. We leverage this high-throughput capability to evaluate R$^{2}$E across a diverse suite of benchmarks, ranging from agile locomotion to complex manipulation, which utilize the specific observation and action spaces defined in Table~\ref{tab:isaacgym_specs}.

{
    \setlength{\textfloatsep}{6pt plus 2pt minus 2pt} 
    \setlength{\intextsep}{6pt plus 2pt minus 2pt}    

    \begin{figure}[htbp]
        \centering
        \includegraphics[width=0.85\linewidth]{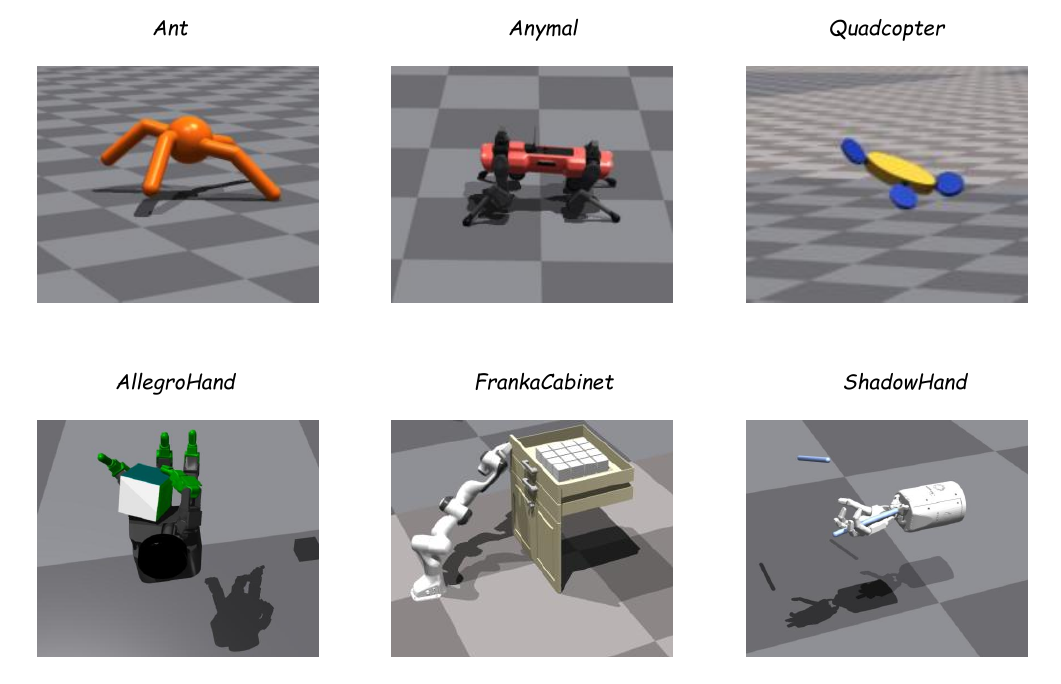} 

\caption{\textbf{Visualization of IsaacGym Tasks.} The benchmark suite is categorized into locomotion tasks (\emph{Ant}, \emph{Anymal}, \emph{Quadcopter}) and manipulation tasks (\emph{AllegroHand}, \emph{FrankaCabinet}, \emph{ShadowHand}).}
        \label{fig:isaacgym_viz}
    \end{figure}

    \begin{table}[htbp]
        \centering
        \caption{\textbf{Specifications of IsaacGym Environments.} Summary of observation (Obs.) and action (Act.) dimensions along with primary task objectives.}

        \label{tab:isaacgym_specs}
        \renewcommand{\arraystretch}{1.1} 
        \small 
        \begin{tabular}{@{}lccp{8.0cm}@{}}
            \toprule
            \textbf{Task} & \textbf{Obs.} & \textbf{Act.} & \textbf{Task Description} \\ 
            \midrule
            \emph{Ant} & 60 & 8 & Maximize forward velocity while maintaining stability. \\ 
            \emph{Anymal} & 48 & 12 & Track random linear and angular velocity commands. \\ 
            \emph{Quadcopter} & 21 & 12 & Navigate to target coordinates and maintain stable hover. \\ 
            \emph{AllegroHand} & 88 & 16 & Reorient object to target pose using a 16-DoF hand. \\ 
            \emph{FrankaCabinet} & 23 & 9 & Interact with and open a cabinet drawer. \\ 
            \emph{ShadowHand} & 211 & 20 & Reorient object using a 24-DoF anthropomorphic hand. \\ 
            \bottomrule
        \end{tabular}
    \end{table}
}

\subsection{Bi-DexHands}

\begin{figure}[H]
    \centering
    \includegraphics[width=0.95\linewidth]{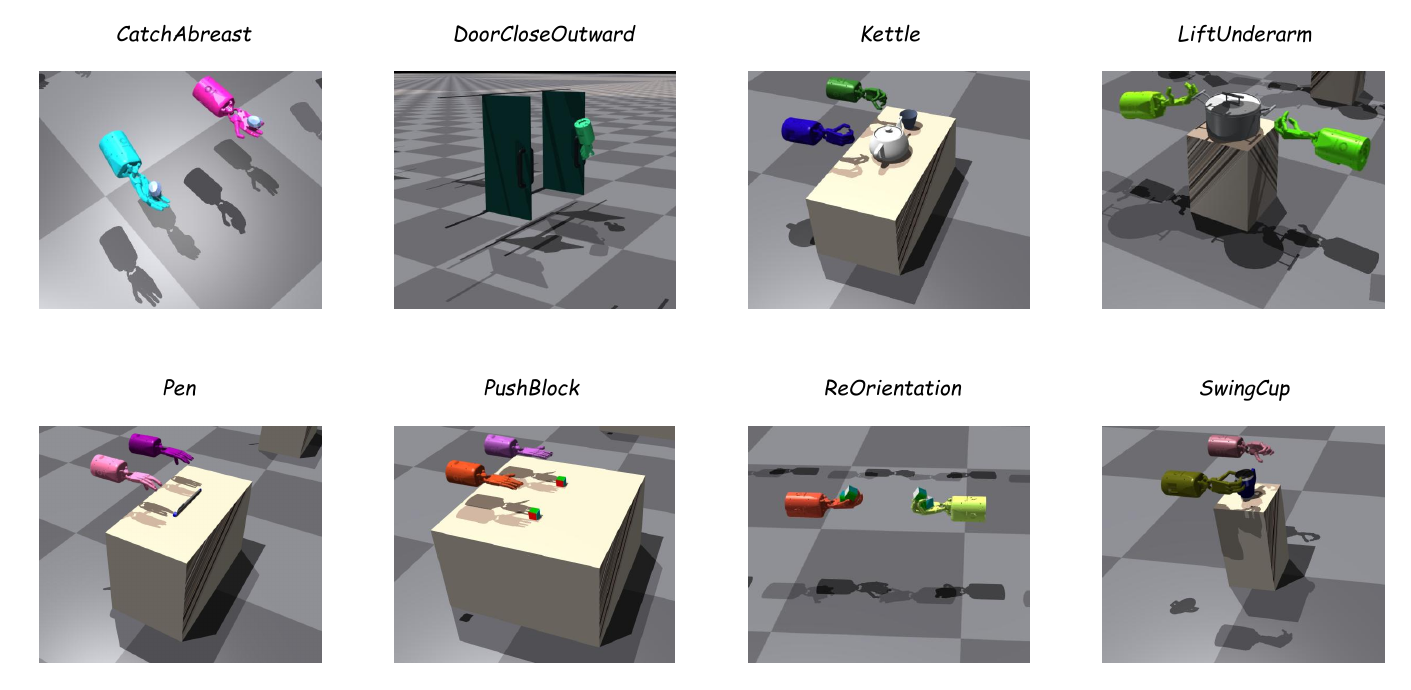}
    \caption{\textbf{Visualization of Bi-DexHands Tasks.} These tasks require synchronized control of two agents to solve high-dimensional manipulation problems, such as pouring water (\emph{Kettle}) or collaborative swinging (\emph{SwingCup}).}
    \label{fig:bidexhands_viz}
\end{figure}

Bi-DexHands~\cite{chen2023bi} represents a significant escalation in control complexity, offering a comprehensive library of bimanual dexterous manipulation tasks built upon the high-performance IsaacGym simulator. Designed to match the dexterity of human bimanual skills, this benchmark necessitates the precise coordination of two multi-joint Shadow Hands to solve intricate cooperation challenges in a shared workspace. As illustrated in Figure~\ref{fig:bidexhands_viz}, the selected environments span a wide spectrum of contact-rich scenarios—ranging from collaborative object handover to coordinated lifting and throwing—thereby serving as a rigorous testbed for evaluating policy synthesis in high-dimensional, multi-agent continuous control spaces.

\renewcommand{\arraystretch}{1.2}
\small

\setlength{\LTcapwidth}{\textwidth}
\begin{longtable}{@{}lccp{8.5cm}@{}}

    \caption{\textbf{Detailed Specifications of Bi-DexHands Environments.} The table lists the high-dimensional observation (Obs.) and action (Act.) spaces alongside the specific coordination logic for each task.} \label{tab:bidexhands_specs} \\
    
    \toprule
    \textbf{Task} & \textbf{Obs.} & \textbf{Act.} & \textbf{Task Description} \\ \midrule
    \endfirsthead
    
    \multicolumn{4}{c}{{\bfseries \tablename\ \thetable{} -- continued from previous page}} \\
    \toprule
    \textbf{Task} & \textbf{Obs.} & \textbf{Act.} & \textbf{Task Description} \\ \midrule
    \endhead
    
    \bottomrule
    \endfoot

    \emph{CatchAbreast} & 422 & 52 & 
    This class corresponds to the Catch Abreast task. This environment consists
      of two shadow hands placed  side by side in the same direction and an object that
      needs to be passed. Compared with the previous  environment which is more like passing
      objects between the hands of two people, this environment is  designed to simulate
      the two hands of the same person passing objects, so different catch techniques  are
      also required and require more hand translation and rotation techniques \\ \midrule
    
    \emph{DoorCloseOutward} & 417 & 52 & 
    This class corresponds to the DoorCloseOutward task. This environment
      also require a closed door  to be opened and the door can only be pushed inward
      or initially open outward, but because they  can't complete the task by simply pushing,
      which need to catch the handle by hand and then open  or close it, so it is relatively
      difficult \\ \midrule
    
    \emph{Kettle} & 417 & 52 & 
    This class corresponds to the PourWater task. This environment involves
      two hands, a kettle, and a bucket, we need to hold the kettle with one hand and the bucket
      with the other hand, and pour the water from the kettle into the bucket. In the
      practice task in IsaacGym, we use many small balls to simulate the water \\ \midrule
    
    \emph{LiftUnderarm} & 417 & 52 & 
    This class corresponds to the LiftUnderarm task. This environment requires
      grasping the pot handle  with two hands and lifting the pot to the designated position.
      This environment is designed to  simulate the scene of lift in daily life and is
      a practical skill\\ \midrule

    \emph{Pen} & 417 & 52 & 
    This class corresponds to the Open Pen Cap task. This environment involves
      two hands and a pen, we need to use two hands to open the pen cap \\ \midrule
    
    \emph{PushBlock} & 428 & 52 & 
    This class corresponds to the PushBlock task. This environment involves
      two hands and two blocks,  we need to use both hands to reach and push the block
      to the desired goal separately. This is a  relatively simple task \\ \midrule
    
    \emph{ReOrientation} & 422 & 40 & 
    This class corresponds to the ReOrientation task. This environment involves
      two hands and two objects.  Each hand holds an object and we need to reorient the
      object to the target orientation \\ \midrule
    
    \emph{SwingCup} & 417 & 52 & 
    This class corresponds to the SwingCup task. This environment involves
      two hands and a dual handle  cup, we need to use two hands to hold and swing the
      cup together \\ 

\end{longtable}

\FloatBarrier

\section{Implementation Details}
\label{app:implementation_details}
This section provides the comprehensive experimental configurations required for reproducibility, covering the massively parallel simulation settings, LLM sampling strategies, and the robust code execution pipeline.

\subsection{RL Training Configuration}

We utilize the Proximal Policy Optimization (PPO) algorithm~\cite{makoviichuk2021rl} as the core policy optimizer. To ensure rigorous reproducibility and isolate the contribution of the evolved reward functions, we strictly adhere to the default hyperparameter configurations (e.g., network architecture, learning rate, clipping range) provided in the official IsaacGym and Bi-DexHands repositories. Consequently, no task-specific tuning was performed on the PPO hyperparameters.

The primary computational variable adjusted across tasks is the number of parallel environments ($N_{\text{envs}}$). This parameter is optimized to fully saturate the GPU memory and maximize simulation throughput, thereby accelerating the fitness evaluation of reward candidates. The specific environment counts for all evaluated tasks are enumerated in Table~\ref{tab:parallel_envs}.

\begin{table}[h!]
    \centering
    \caption{\textbf{Parallel Environment Configurations.} The specific number of parallel simulation instances ($N_{\text{envs}}$) utilized for each task to maximize GPU throughput.}
    \label{tab:parallel_envs}
    \renewcommand{\arraystretch}{1.1}
    \small
    \begin{tabular}{@{}lc|lc@{}}
    \toprule
    \multicolumn{2}{c|}{\textbf{IsaacGym Tasks}} & \multicolumn{2}{c}{\textbf{Bi-DexHands Tasks}} \\ 
    \cmidrule(r){1-2} \cmidrule(l){3-4}
    \textbf{Task Name} & \textbf{$N_{\text{envs}}$} & \textbf{Task Name} & \textbf{$N_{\text{envs}}$} \\ 
    \midrule
    AllegroHand & 16,384 & CatchAbreast & 2,048 \\
    Quadcopter & 8,192 & DoorCloseOutward & 2,048 \\
    ShadowHand & 8,192 & Kettle & 2,048 \\
    Ant & 4,096 & LiftUnderarm & 2,048 \\
    Anymal & 4,096 & Pen & 2,048 \\
    FrankaCabinet & 4,096 & PushBlock & 2,048 \\
    & & ReOrientation & 2,048 \\
    & & SwingCup & 2,048 \\
    \bottomrule
    \end{tabular}
\end{table}

\subsection{LLM Generation \& Evolutionary Settings}
\label{app:llm_settings}

To ensure full reproducibility and facilitate future research, we detail the stochastic sampling configurations and the evolutionary hyperparameters used to instantiate the R$^{2}$E framework.

\paragraph{Sampling Parameters}
We leverage the \emph{Qwen3-Max} model via its high-throughput completion API. The sampling strategy is designed to balance the generation of syntactically correct code with the creativity required for reward discovery:
\begin{itemize}
    \setlength{\itemsep}{2pt} 
    \item \textbf{Temperature ($T_{llm} = 1.0$):} We maintain a constant temperature of 1.0 across all roles (\emph{Explorer}, \emph{Guardian}, \emph{Evolver}). Unlike standard code generation tasks that prefer lower temperatures for precision, reward synthesis requires traversing a non-convex optimization landscape. This higher entropy setting prevents the model from converging prematurely to conventional coding patterns, ensuring sufficient diversity in the structural logic and coefficient selection.
    
    \item \textbf{Batch Sampling ($n$):} To minimize wall-clock latency, we utilize the \emph{n} parameter to trigger parallel batch sampling. Each role's assigned budget is generated within a single request context, effectively utilizing the provider's limits (600 RPM / 1M TPM). This approach reduces the overhead associated with establishing multiple HTTP connections.
    
    \item \textbf{Max Output Tokens (64k):} The output limit is set to 64,000 tokens. This generous upper bound is critical for the \emph{Evolver} role, which often engages in "Chain-of-Thought" reasoning before outputting the final code. It ensures that complex reward functions, which may include extensive logic branches and internal state tracking, are never truncated mid-generation.
    
    \item \textbf{Context Window (256k):} We utilize the full 256k context window to implement a "long-term memory" mechanism. This allows the prompt to include a comprehensive history of the \textit{Global Elite Pool}, detailed error logs from failed executions, and multiple high-performing reference functions from previous iterations without exceeding input constraints.
\end{itemize}

\paragraph{Evolutionary Search Configuration}
The evolutionary process is governed by a set of structural constants empirically selected to balance computational frugality with search depth:
\begin{itemize}
    \setlength{\itemsep}{2pt}
    \item \textbf{Sample Budget ($N=16$):} We generate 16 candidate reward functions per iteration. This number provides a statistically significant sample size to cover different mutation directions while keeping the simulation cost manageable.
    \item \textbf{Search Horizon ($T=5$):} The optimization runs for 5 iterative rounds. Our preliminary experiments indicated that marginal gains diminish significantly after the 5th iteration, making this a cost-effective stopping criterion.
    \item \textbf{Global Elite Pool Size ($K=5$):} The Top-5 performing rewards are retained in the Elite Pool. This strategy prevents catastrophic forgetting by ensuring that the best genes are always available for crossover in subsequent generations.
\end{itemize}

\paragraph{Dynamic Role Scheduling Strategy}
From the second iteration ($t > 1$) onwards, the total sample budget $N=16$ is dynamically partitioned among the specialized roles. This partition follows a linear decay schedule  designed to mimic a \textit{Simulated Annealing} process. Let $t \in \{2, \dots, T\}$ be the current iteration index. The allocation is calculated as follows:

\begin{align}
    N_{\text{Explorer}}(t) &= \left\lceil N \times \left( 1 - \frac{t-1}{T-1} \right) \right\rceil \\
    N_{\text{Guardian}}(t) &= \left\lfloor \frac{N - N_{\text{Explorer}}(t)}{2} \right\rfloor \\
    N_{\text{Evolver}}(t) &= N - N_{\text{Explorer}}(t) - N_{\text{Guardian}}(t)
\end{align}

\noindent \textbf{Theoretical Interpretation:}
This scheduling strategy facilitates a smooth transition from \textit{Exploration} to \textit{Exploitation}. In the early stages (e.g., $t=2$), $N_{\text{Explorer}}$ dominates, encouraging the framework to cast a wide net and discover novel reward structures. As $t$ approaches $T$, the budget shifts towards the \emph{Guardian} and \emph{Evolver}, focusing resources on refining the numerical stability of existing elites and performing local crossover operations. At $t=1$, to maximize initial coverage, all $N$ samples are generated via a role-agnostic initialization prompt.

\section{Additional Experimental Results}
\label{app:additional_experimental_results}

\subsection{Evolution of the Best-So-Far Performance}
\label{appendix:max_success_rate}

To provide a more granular analysis of the reward search efficiency, we track the best performance discovered up to iteration $t$, averaged over 5 random seeds, across 14 tasks from IsaacGym and Bi-DexHands. Unlike the learning curve of the reward evaluated at the current iteration, this quantity records the best reward found so far. On IsaacGym it is the task-specific performance metric; on Bi-DexHands it is the success rate. It serves as a direct proxy for optimization efficiency and for the ability to escape local optima.

\begin{figure}[H] 
    \centering
    \includegraphics[width=\linewidth]{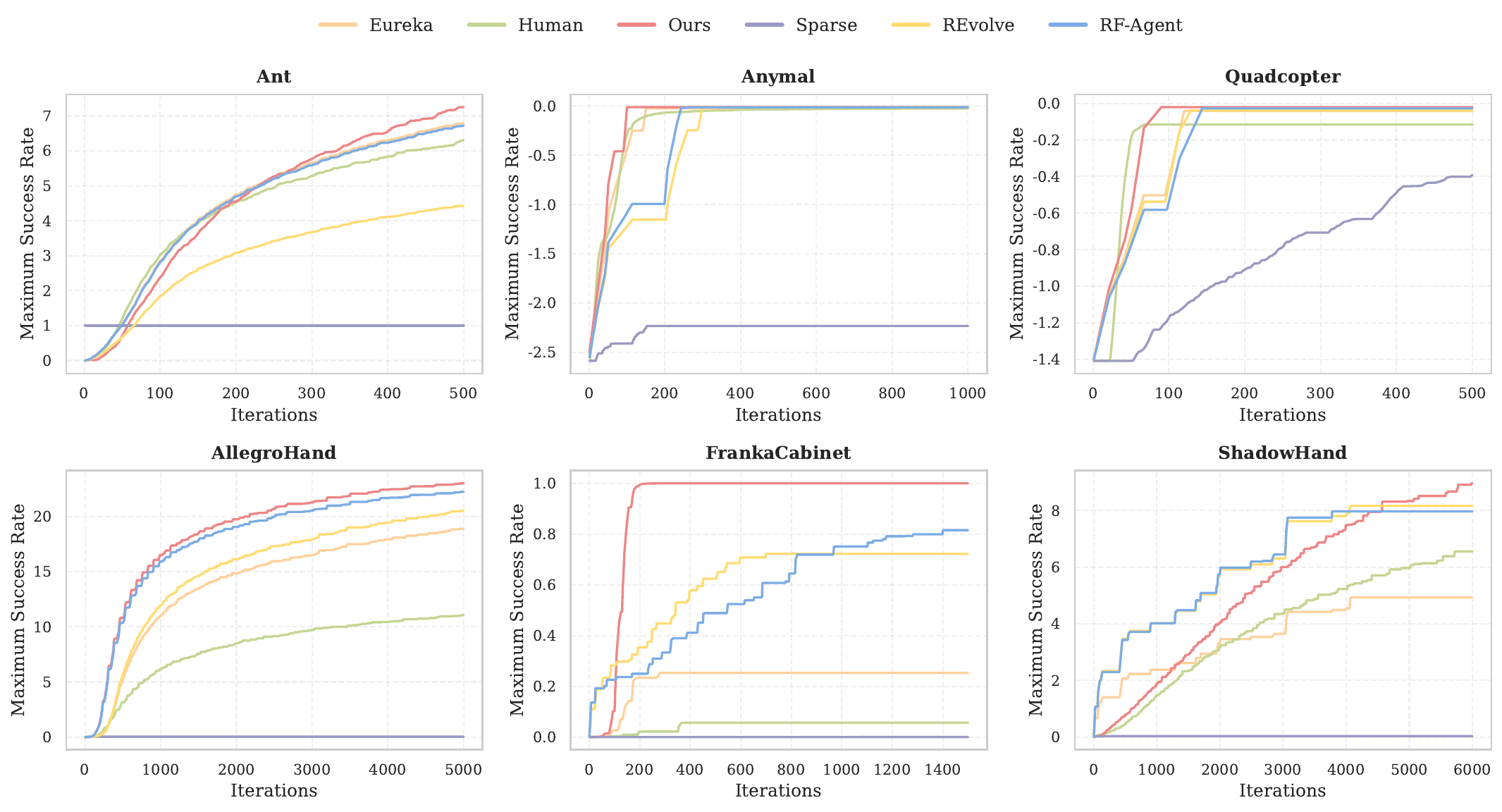}
    \caption{\textbf{Evolution of the best task-specific performance metric on IsaacGym.} The curves represent the highest task-specific performance metric achieved by the best reward discovered so far as the number of iterations increases. Our method (R$^{2}$E) demonstrates faster convergence (e.g., \emph{FrankaCabinet}) and higher asymptotic performance (e.g., \emph{AllegroHand}, \emph{ShadowHand}) compared to baselines.}
    \label{fig:max_isaacgym}

\end{figure}

\begin{figure}[H]
    \centering
    \includegraphics[width=\linewidth]{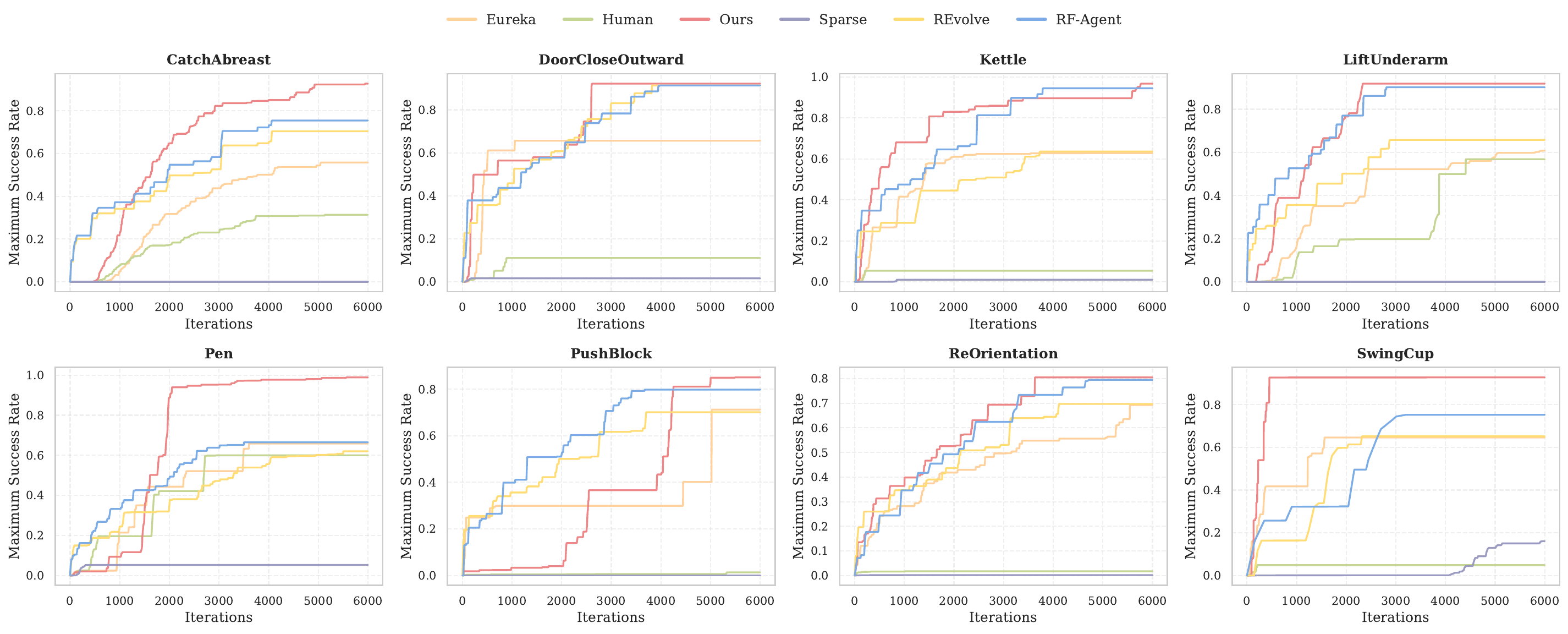}
    \caption{\textbf{Evolution of Maximum Success Rate on Bi-DexHands Tasks.} Comparisons across 8 complex manipulation tasks show that R$^{2}$E consistently locates superior reward functions earlier in the search process. Notably, in difficult tasks like \emph{Pen}, R$^{2}$E succeeds in finding functional rewards where baselines stagnate.}
    \label{fig:max_bidexhands}
\end{figure}

The results for IsaacGym and Bi-DexHands are illustrated in Figure~\ref{fig:max_isaacgym} and Figure~\ref{fig:max_bidexhands}, respectively.

\paragraph{Convergence Speed.}
As evident from the figures, R$^{2}$E exhibits significantly faster convergence rates compared to the baselines across the majority of tasks. In high-dimensional environments, our method rapidly identifies high-quality reward regions within the first few iterations. For instance, in \emph{FrankaCabinet} (Fig.~\ref{fig:max_isaacgym}), R$^{2}$E achieves a steep performance jump almost immediately, whereas baselines requires significantly more iterations to reach comparable levels. Similarly, in \emph{SwingCup} (Fig.~\ref{fig:max_bidexhands}), our method establishes a clear lead early in the search. This accelerated discovery is attributed to the Explorer role, which actively diversifies the search space, and the Global Elite Pool, which ensures that promising structures are immediately preserved and used to seed subsequent generations.

\paragraph{Asymptotic Performance and Local Optima Escaping.}
In addition to search efficiency, R$^{2}$E reaches a higher final task-specific performance metric on IsaacGym and a higher final success rate on Bi-DexHands. In tasks where baselines plateau early—indicating convergence to local optima—our approach continues to improve. This is most strikingly observed in \emph{CatchAbreast} (Fig.~\ref{fig:max_bidexhands}), where baselines fail to make meaningful progress, while R$^{2}$E successfully discovers reward functions that solve the task. Furthermore, in fine-grained manipulation tasks such as \emph{PushBlock} (Fig.~\ref{fig:max_bidexhands}) and \emph{Kettle} (Fig.~\ref{fig:max_bidexhands}), R$^{2}$E widens the performance gap in later iterations. This sustained improvement demonstrates the effectiveness of the Guardian and Evolver roles, which continuously refine and recombine elite traits to push the performance boundary beyond what standard iterative prompting can achieve.

\FloatBarrier 

\subsection{Computational Cost Analysis}
\label{app:cost_analysis}

The computational efficiency of R$^{2}$E is primarily determined by the parallelization capability of the RL training stage within IsaacGym. While the LLM generation phase (via \emph{Qwen3-Max}) is relatively constant (taking approximately 1--3 minutes per iteration), the RL training time scales with the complexity of the environment's physics and the dimensionality of the action space.

\paragraph{Total Wall-clock Time}
Based on our experiments conducted on a 4$\times$ \emph{NVIDIA RTX 4090} setup, the total time required to complete a full evolutionary search (5 iterations, $N=16$ candidates per iteration) varies across tasks. We categorize the empirical execution times as follows:

\begin{itemize}
    \item \textbf{Low Complexity (Locomotion/Control):} Tasks such as \emph{Ant} and \emph{Quadcopter} converge very rapidly, completing the entire 5-iteration search in approximately \textbf{45--50 minutes}.
    \item \textbf{Medium Complexity (Basic Manipulation):} Tasks involving robotic arms or basic legged locomotion, such as \emph{FrankaCabinet} and \emph{Anymal}, require approximately \textbf{1.5--2 hours}.
    \item \textbf{High Complexity (Dexterous Manipulation):} Intricate tasks using the Shadow Hand or Allegro Hand (e.g., \emph{Pen}, \emph{AllegroHand}) exhibit the highest computational demand due to complex contact physics, with total search times ranging from \textbf{6 to 12 hours}.
\end{itemize}

\begin{table}[H] 
    \centering
    \caption{Empirical wall-clock time for representative tasks (Total 5-iteration search).}
    \label{tab:detailed_time}
    \renewcommand{\arraystretch}{1.1} 
    \small
    \begin{tabular}{lc}
        \toprule 
        \textbf{Task Name} & \textbf{Total Time (H:M:S)} \\
        \midrule
        \emph{AllegroHand}        & 11:57:05 \\
        \emph{Ant}                & 00:48:19 \\
        \emph{Anymal}             & 01:46:43 \\
        \emph{FrankaCabinet}      & 01:34:50 \\
        \emph{Quadcopter}         & 00:46:03 \\
        \emph{ShadowHand}         & 06:42:25 \\
        \midrule
        \emph{Catch Abreast}      & 06:16:30 \\
        \emph{Door Close Outward} & 08:58:32 \\
        \emph{Kettle}             & 12:09:01 \\
        \emph{Lift Underarm}      & 08:18:31 \\
        \emph{Pen}                & 07:55:03 \\
        \emph{Push Block}         & 05:04:10 \\
        \emph{Re-orientation}     & 05:46:50 \\
        \emph{Swing Cup}          & 07:38:13 \\
        \bottomrule
    \end{tabular}
\end{table}

\paragraph{Analysis of Search Efficiency}
The results indicate that for simpler tasks, R$^{2}$E can discover near-optimal reward functions in under an hour. Even for the most challenging dexterous manipulation tasks in Bi-DexHands, the search concludes within half a day. Given that each iteration involves training 16 distinct policies, the average training time per candidate remains highly efficient due to our parallelized infrastructure. This demonstrates that R$^{2}$E is a practical and scalable reward for automated reward engineering in complex robotic domains.

\FloatBarrier 

\subsection{Token Usage and Economic Cost Analysis}
\label{app:token_cost}

To rigorously evaluate the economic feasibility of deploying R$^{2}$E in large-scale experiments, we implemented a granular monitoring system to track the token consumption and associated API costs for each distinct task. Our cost analysis is based on the \emph{Qwen3-Max} model utilizing the Batch Inference pricing tier. This tier is particularly suitable for evolutionary search processes where real-time latency is not a critical constraint, allowing for significant cost reductions compared to standard synchronous API calls.

\paragraph{Pricing Model and Accounting Method}
The reported costs are calculated based on the official batch inference rates for the China region (RMB), which are structured as follows:
\begin{itemize}
    \item \textbf{Input Processing (Batch):} 0.0016 RMB per 1,000 tokens.
    \item \textbf{Output Generation (Batch):} 0.0064 RMB per 1,000 tokens.
\end{itemize}
The cumulative token count presented includes all prompt inputs (System, User, and Assistant context) and generated completions across all three evolutionary roles (\emph{Explorer}, \emph{Guardian}, \emph{Evolver}) over the entire course of the 5-iteration search process. This ensures that the cost estimates reflect the complete end-to-end expense of synthesizing a finalized reward function.

\paragraph{Analysis of Consumption Patterns}
Table~\ref{tab:token_usage} summarizes the detailed token usage and estimated total cost. We observe a distinct correlation between task complexity and token consumption. Standard locomotion and single-arm manipulation tasks (e.g., \emph{Ant}, \emph{Quadcopter}, \emph{FrankaCabinet}) typically consume between 90k to 120k total tokens. In contrast, high-dimensional bimanual manipulation tasks from the Bi-DexHands benchmark (e.g., \emph{DoorCloseOutward}, \emph{SwingCup}) exhibit a marked increase in usage, ranging from 170k to over 210k tokens. This approximate $2\times$ increase is attributed to the verbose context required to describe the dual-arm state spaces (often exceeding 200 dimensions) and the syntactic complexity involved in generating coordinated reward logic for multi-agent-like interactions.

\begin{table}[H]
    \centering
    \caption{\textbf{Token Usage and Economic Cost Analysis.} The table details the cumulative prompt and completion tokens consumed over a full 5-iteration evolutionary search, along with the estimated total API cost in RMB.}
    \label{tab:token_usage}
    \renewcommand{\arraystretch}{1.1} 
    \setlength{\tabcolsep}{4pt}
    \begin{small}
    \resizebox{0.85\textwidth}{!}{
    \begin{tabular}{lcccc}
        \toprule
        \textbf{Task Name} & \textbf{Prompt Tokens} & \textbf{Completion Tokens} & \textbf{Total Tokens} & \textbf{Est. Cost (RMB)} \\
        \midrule
        \emph{AllegroHand} & 76,118 & 31,464 & 107,582 & 0.32 \\
        \emph{Ant} & 73,182 & 38,234 & 111,416 & 0.36 \\
        \emph{Anymal} & 69,530 & 48,141 & 117,671 & 0.42 \\
        \emph{FrankaCabinet} & 60,717 & 34,121 & 94,838 & 0.32 \\
        \emph{Quadcopter} & 56,514 & 35,221 & 91,735 & 0.32 \\
        \emph{ShadowHand} & 75,419 & 23,899 & 99,318 & 0.27 \\
        \emph{CatchAbreast} & 66,860 & 49,772 & 116,632 & 0.43 \\
        \emph{DoorCloseOutward} & 127,790 & 73,231 & 201,021 & 0.67 \\
        \emph{Kettle} & 117,016 & 56,974 & 173,990 & 0.55 \\
        \emph{LiftUnderarm} & 83,272 & 48,480 & 131,752 & 0.44 \\
        \emph{Pen} & 124,741 & 72,784 & 197,525 & 0.67 \\
        \emph{PushBlock} & 114,172 & 49,836 & 164,008 & 0.50 \\
        \emph{ReOrientation} & 84,855 & 59,122 & 143,977 & 0.51 \\
        \emph{SwingCup} & 128,453 & 83,361 & 211,814 & 0.74 \\
        \bottomrule
    \end{tabular}
    }
    \end{small}
\end{table}

\paragraph{Economic Efficiency and Scalability}
The financial implications of R$^{2}$E are minimal when compared to traditional reward engineering workflows. As evidenced in Table~\ref{tab:token_usage}, the total API cost for synthesizing a high-performing reward function remains under \textbf{1.0 RMB} (approximately \$0.14 USD) for every evaluated task. Even for the most intricate task, \emph{SwingCup}, the cost caps at merely 0.74 RMB. 

To put this into perspective, the computational cost of training the RL agent itself (GPU electricity and amortization) typically exceeds the LLM query cost by several orders of magnitude. Furthermore, compared to the labor cost of a domain expert—who might spend days iterating on reward coefficients—R$^{2}$E offers a cost reduction of several orders of magnitude. This extreme economic efficiency demonstrates that the R$^{2}$E framework is highly scalable and can be deployed to solve hundreds of diverse robotic tasks without incurring significant financial overhead.

\subsection{Effect of LLM Backbone}
\label{app:llm_backbone}

We examine the sensitivity of reward evolution to different LLM backbones by substituting the generator used in both Eureka and R$^{2}$E, while keeping the same environments and evaluation protocol. Table~\ref{tab:llm_backbone} reports mean $\pm$ standard deviation over 5 seeds for two representative tasks.

\begin{table}[H]
    \centering
    \caption{\textbf{Impact of LLM Backbone on Reward Evolution.} Performance is reported as mean $\pm$ standard deviation over 5 seeds.}
    \label{tab:llm_backbone}
    \renewcommand{\arraystretch}{1.1}
    \setlength{\tabcolsep}{6pt}
    \begin{small}
    \begin{tabular}{l l c c}
        \toprule
        \textbf{Task} & \textbf{LLM Backbone} & \textbf{Eureka} & \textbf{R$^{2}$E} \\
        \midrule
        \textbf{Ant} & Qwen3-Coder-Flash & 3.8851{\scriptsize $\pm$0.0289} & 6.2114{\scriptsize $\pm$0.6671} \\
        & Qwen3-Max & 6.5876{\scriptsize $\pm$0.2602} & 7.2639{\scriptsize $\pm$0.4584} \\
        & GPT-5.4 & 6.3583{\scriptsize $\pm$0.6967} & 7.2615{\scriptsize $\pm$0.5690} \\
        \textbf{ShadowHand} & Qwen3-Coder-Flash & 0.1626{\scriptsize $\pm$0.0866} & 0.5577{\scriptsize $\pm$0.3162} \\
        & Qwen3-Max & 4.9272{\scriptsize $\pm$1.4171} & 8.9603{\scriptsize $\pm$1.4174} \\
        & GPT-5.4 & 7.8988{\scriptsize $\pm$0.5859} & 9.6329{\scriptsize $\pm$1.3460} \\
        \bottomrule
    \end{tabular}
    \end{small}
\end{table}

From the results, we observe a clear trend that stronger LLM backbones consistently lead to improved performance. Notably, R$^{2}$E benefits more significantly from increased model capability, yielding larger performance gains as the backbone becomes more powerful. We attribute this to the fact that stronger models exhibit a better understanding of role specification and instruction following, which is crucial for R$^{2}$E's role-conditioned reward evolution process. This enhanced role awareness enables more coherent and effective reward design, ultimately leading to superior downstream policy performance.

\subsection{LLM Backbone Diversity and Reward Homogenization}
\label{app:backbone_diversity}

Different backbone models can follow different reasoning patterns, possibly because post-training imposes different optimization objectives and alignment strategies. To examine this difference, we compare \emph{Qwen3.7-Max}, \emph{DeepSeek-v4-pro}, and \emph{GPT-5.6-sol-max} on \emph{Ant} and \emph{Pen}.

The three models approach reward design differently.
\emph{Qwen3.7-Max} first uses reliable signals already available in the environment and then adds reward terms based on those signals.
\emph{DeepSeek-v4-pro} first decomposes the task into intuitively reasonable behaviors and then combines the corresponding reward terms.
\emph{GPT-5.6-sol-max} first considers which reward designs are likely to cause training failures and then infers which combinations of reward terms are better aligned with the task.
These observations suggest that using different models can reduce homogeneity among the sampled reward functions. They do not imply that switching the backbone across iterations within a single run will improve the final task performance.

Table~\ref{tab:mixed_backbone} reports the final evaluation. \emph{Ant} is measured by its task-specific performance metric, and \emph{Pen} by success rate. In the mixed setting, the backbone is assigned by iteration: iterations 1 through 5 use \emph{Qwen3.7-Max}, \emph{DeepSeek-v4-pro}, \emph{GPT-5.6-sol-max}, \emph{Qwen3.7-Max}, and \emph{DeepSeek-v4-pro}, respectively. Alternating among the three models does not outperform the strongest single backbone on either task. We conjecture that switching models can disrupt the coherence of the reasoning process across iterations.

\begin{table}[H]
    \centering
    \caption{\textbf{Single and mixed LLM backbones.} \emph{Ant} reports the task-specific performance metric, and \emph{Pen} reports success rate. Results are mean $\pm$ standard deviation.}
    \label{tab:mixed_backbone}
    \renewcommand{\arraystretch}{1.1}
    \setlength{\tabcolsep}{6pt}
    \begin{small}
    \begin{tabular}{lcc}
        \toprule
        \textbf{LLM Backbone} & \textbf{Ant} & \textbf{Pen} \\
        \midrule
        Qwen3.7-Max only & $7.5604 \pm 0.6164$ & $0.9678 \pm 0.0057$ \\
        DeepSeek-v4-pro only & $7.6408 \pm 0.1465$ & $0.9468 \pm 0.0416$ \\
        GPT-5.6-sol-max only & $7.9866 \pm 0.3364$ & $0.9769 \pm 0.0014$ \\
        Mixed (alternating LLMs) & $7.7542 \pm 0.2541$ & $0.9538 \pm 0.0048$ \\
        \bottomrule
    \end{tabular}
    \end{small}
\end{table}

\section{Limitation}
\label{app:E}
\textbf{Limitation.} \quad
Our R$^{2}$E method, like Eureka, REvolve, and RF-Agent, falls into the category of reward function design approaches that depend on iterative feedback from LLMs combined with repeated policy optimization. Consequently, the overall pipeline incurs substantial computational cost and time overhead, as each task involves multiple rounds of LLM interaction alongside several reinforcement learning training cycles. Although R$^{2}$E can deliver strong generation performance even when built on relatively small LLM backbones, it does not alleviate the need for numerous RL iterations in the outer-loop search. In future work, we aim to reduce the number of required RL training cycles while preserving the advantages of iterative reward refinement.

A further limitation of our approach is its reliance on a sufficiently strong base model. Since the method depends on accurate role understanding and nuanced feedback, weaker backbone models tend to exhibit degraded performance, which in turn negatively impacts the overall effectiveness of the system.

\section{Algorithm}

The operational workflow of R$^{2}$E is summarized in Algorithm~\ref{alg:r2e}. 
It begins by generating an initial set of reward functions to explore the reward space and establish a diverse seed population. 
From the second iteration onward, candidates are evolved through the Explorer, Guardian, and Evolver roles under a dynamic scheduling strategy that shifts focus from exploration to refinement. 
Each candidate is evaluated via RL training, and best-performing rewards are stored in the Global Elite Pool to guide subsequent search. 
This process balances novelty and stability, mitigating instability and enabling effective evolution of reward functions.

\begin{algorithm}[H]
\caption{\textbf{R$^{2}$E}: Role-driven Reward Evolutionary Framework}
\label{alg:r2e}
\begin{algorithmic}[1]
\REQUIRE Environment $E$, LLM $M$, total iterations $T$, budget per iteration $N$, pool size $K$
\STATE Initialize Global Elite Pool $\mathcal{B} \leftarrow \emptyset$
\FOR{$t = 1$ \textbf{to} $T$}
    \IF{$t == 1$}
        \STATE Generate $N$ initial reward functions $\{r_i\}_{i=1}^N$ using $M$ to broadly explore the reward space
    \ELSE
        \STATE Determine role allocation $\{N_{\text{Exp}}, N_{\text{Gua}}, N_{\text{Evo}}\}$ via dynamic scheduling strategy
        \STATE \textbf{Explorer}: Query $r_{best}$ and $r_{rand} \in \mathcal{B}$ to generate $N_{\text{Exp}}$ novel candidates
        \STATE \textbf{Guardian}: Query Top-1 reward $r_{best} \in \mathcal{B}$ to generate $N_{\text{Gua}}$ stable variations
        \STATE \textbf{Evolver}: Query $r_{best}$ and $r_{rand} \in \mathcal{B}$ to cross over and mutate $N_{\text{Evo}}$ offspring
        \STATE Aggregate all generated reward functions $\{r_i\}_{i=1}^N$
    \ENDIF
    \FOR{each reward function $r_i$}
        \STATE Execute RL training process in $E$ guided by $r_i$
        \STATE Evaluate task performance $\mathcal{J}(\pi_{r_i})$ via validation rollouts
        \IF{$|\mathcal{B}| < K$}
            \STATE Insert $r_i$ into $\mathcal{B}$
        \ELSIF{$\mathcal{J}(\pi_{r_i}) > \min_{r \in \mathcal{B}} \mathcal{J}(\pi_ r)$}
            \STATE Replace the lowest performing reward in $\mathcal{B}$ with $r_i$
        \ENDIF
    \ENDFOR
\ENDFOR
\STATE \textbf{Return} the top-performing reward function $r^\ast \in \mathcal{B}$
\end{algorithmic}
\end{algorithm}

\newpage
\section*{NeurIPS Paper Checklist}

\begin{enumerate}

\item {\bf Claims}
    \item[] Question: Do the main claims made in the abstract and introduction accurately reflect the paper's contributions and scope?
    \item[] Answer: \answerYes{} 
    \item[] Justification: We accurately claim the main contributions and scope of our paper in the Abstract and Introduction section.

\item {\bf Limitations}
    \item[] Question: Does the paper discuss the limitations of the work performed by the authors?
    \item[] Answer: \answerYes{} 
    \item[] Justification: We discuss the limitations of our work in Appendix~\ref{app:E}.

\item {\bf Theory assumptions and proofs}
    \item[] Question: For each theoretical result, does the paper provide the full set of assumptions and a complete (and correct) proof?
    \item[] Answer: \answerNA{} 
    \item[] Justification: Our paper does not include theoretical results, so this question is not applicable to our paper.

    \item {\bf Experimental result reproducibility}
    \item[] Question: Does the paper fully disclose all the information needed to reproduce the main experimental results of the paper to the extent that it affects the main claims and/or conclusions of the paper (regardless of whether the code and data are provided or not)?
    \item[] Answer: \answerYes{} 
    \item[] Justification: Appendix \ref{app:implementation_details} contains all information required to reproduce our experimental results. Note that due to the inherent randomness of LLMs, exact numerical values cannot be perfectly reproduced, while the overall result range and comparative trends are reproducible.

\item {\bf Open access to data and code}
    \item[] Question: Does the paper provide open access to the data and code, with sufficient instructions to faithfully reproduce the main experimental results, as described in supplemental material?
    \item[] Answer: \answerYes{} 
    \item[] Justification: The full code and experimental data are provided in our supplementary material.

\item {\bf Experimental setting/details}
    \item[] Question: Does the paper specify all the training and test details (e.g., data splits, hyperparameters, how they were chosen, type of optimizer) necessary to understand the results?
    \item[] Answer: \answerYes{} 
    \item[] Justification: Appendix \ref{app:implementation_details} and Section \ref{experimental_setup} provide detailed information about the experimental setup.

\item {\bf Experiment statistical significance}
    \item[] Question: Does the paper report error bars suitably and correctly defined or other appropriate information about the statistical significance of the experiments?
    \item[] Answer: \answerYes{} 
    \item[] Justification: We provide statistical results and illustrate the experimental variability in Section \ref{main_results}.

\item {\bf Experiments compute resources}
    \item[] Question: For each experiment, does the paper provide sufficient information on the computer resources (type of compute workers, memory, time of execution) needed to reproduce the experiments?
    \item[] Answer: \answerYes{} 
    \item[] Justification: We provide the detailed hardware and computing resource information in Appendix \ref{app:cost_analysis}.
    
\item {\bf Code of ethics}
    \item[] Question: Does the research conducted in the paper conform, in every respect, with the NeurIPS Code of Ethics \url{https://neurips.cc/public/EthicsGuidelines}?
    \item[] Answer: \answerYes{} 
    \item[] Justification: This work fully complies with the NeurIPS Code of Ethics.

\item {\bf Broader impacts}
    \item[] Question: Does the paper discuss both potential positive societal impacts and negative societal impacts of the work performed?
    \item[] Answer: \answerNA{} 
    \item[] Justification: This work focuses on reward function design for large language models in embodied scenarios. It belongs to foundational algorithmic research and has no direct negative societal impacts or foreseeable malicious applications.
    
\item {\bf Safeguards}
    \item[] Question: Does the paper describe safeguards that have been put in place for responsible release of data or models that have a high risk for misuse (e.g., pre-trained language models, image generators, or scraped datasets)?
    \item[] Answer: \answerNA{} 
    \item[] Justification: There are no such risks in our paper.

\item {\bf Licenses for existing assets}
    \item[] Question: Are the creators or original owners of assets (e.g., code, data, models), used in the paper, properly credited and are the license and terms of use explicitly mentioned and properly respected?
    \item[] Answer: \answerYes{} 
    \item[] Justification: All publicly available datasets and open-source frameworks used in this work are properly cited, with their original licenses and usage terms respected.

\item {\bf New assets}
    \item[] Question: Are new assets introduced in the paper well documented and is the documentation provided alongside the assets?
    \item[] Answer: \answerYes{} 
    \item[] Justification: Included with the supplementary material.

\item {\bf Crowdsourcing and research with human subjects}
    \item[] Question: For crowdsourcing experiments and research with human subjects, does the paper include the full text of instructions given to participants and screenshots, if applicable, as well as details about compensation (if any)? 
    \item[] Answer: \answerNA{} 
    \item[] Justification: No crowdsourcing or human-subject research is involved in this paper.

\item {\bf Institutional review board (IRB) approvals or equivalent for research with human subjects}
    \item[] Question: Does the paper describe potential risks incurred by study participants, whether such risks were disclosed to the subjects, and whether Institutional Review Board (IRB) approvals (or an equivalent approval/review based on the requirements of your country or institution) were obtained?
    \item[] Answer: \answerNA{} 
    \item[] Justification: This work involves no human-participant experiments, so IRB approval or equivalent ethical review is not required.

\item {\bf Declaration of LLM usage}
    \item[] Question: Does the paper describe the usage of LLMs if it is an important, original, or non-standard component of the core methods in this research? Note that if the LLM is used only for writing, editing, or formatting purposes and does \emph{not} impact the core methodology, scientific rigor, or originality of the research, declaration is not required.
    \item[] Answer: \answerYes{} 
    \item[] Justification: Detailed usage of LLMs is provided in Section \ref{method}, Appendix \ref{app:llm_settings} and Appendix \ref{app:token_cost}.

\end{enumerate}

\end{document}